%% file: neurips_2026.tex
\documentclass{article}

\title{On the Nature of Attention Sink
\\that Shapes Decoding Strategy in Omni-LLMs}

\author{
  Suho Yoo$^{1\:\ast}$ \quad
  Youngjoon Jang$^{1,2}$\thanks{Equal contribution.} \quad
  Joon Son Chung$^1$ \\
  $^1$KAIST\quad $^2$VGG, University of Oxford \\
  \texttt{suho.yoo@kaist.ac.kr} \\[0.5em]
  \href{https://github.com/kaistmm/Outro}{\textcolor{NavyBlue}{https://github.com/kaistmm/Outro}}
}

\input{preamble}

\begin{document}

\maketitle

\begin{abstract}
The goal of this paper is to strengthen the reasoning of Omnimodal Large Language Models (Omni-LLMs) at inference time, without additional training. 
These models jointly process video, audio, and text, and given the large number of tokens they consume, how attention is routed across them is central to their behaviour. 
We focus specifically on \emph{attention sinks}, tokens that absorb a disproportionate share of attention mass regardless of their semantic content, to understand how this routing unfolds. To this end, we conduct a systematic analysis of sink behaviour in Omni-LLMs.
Our analysis yields two key findings: (i)~high sink attention does not solely indicate head redundancy, suggesting that sink value representations play additional functional roles;
(ii)~the sink value vector acts as a shared bias added to every token's output, serving as a global signal that organises the representation as a whole. Building on this, we propose \textit{\textbf{OutRo}}, which correspondingly aligns non-sink token representations with the sink in feature space, and relaxes the causal mask for sink tokens at an early layer to sharpen this bias before the rest of decoding proceeds.
This design enhances the reasoning process without requiring additional forward passes or access to attention maps. Based on extensive experiments, OutRo consistently improves performance on seven video QA benchmarks and demonstrates strong generalisation, while incurring only a $1.1\times$ decoding overhead.

\end{abstract}

\input{sections/1_intro}

\input{sections/5_relatedworks}
\input{sections/2_attentionsink}
\input{sections/3_method}

\input{sections/4_experiments}

\section{Conclusion}
In this work, we presented the first analysis of attention 
sinks in Omni-LLMs and showed that sink tokens are not 
merely structural artifacts: they function as shared bias 
terms in the attention output, injecting a consistent 
direction across all token representations. Building on 
this insight, we proposed \textbf{\textit{OutRo}}, a 
lightweight inference-time method that explicitly aligns 
non-sink token representations with this bias direction 
and sharpens it by relaxing causal masking at sink 
positions. OutRo requires no additional forward passes 
or attention map access, introduces only a $1.1\times$ 
decoding overhead, and consistently improves performance 
across diverse video QA benchmarks. Our findings offer 
a new perspective on how attention sinks shape decoding 
in Omni-LLMs, and suggest that leveraging the sink bias 
is a practical and modality-agnostic strategy for 
inference-time improvement.

\section*{Acknowledgments}
This work was supported by the Institute of Information \& communications Technology Planning \& Evaluation (IITP) through a grant funded by the Korean government (MSIT; RS-2024-00457882, National AI Research Lab Project, 50\%) and by the IITP--ITRC program funded by the Korean government (MSIT; IITP-2026-RS-2023-00259991, 50\%).

\bibliography{shortstrings, main}
\bibliographystyle{plain}

\input{sections/supp}


\end{document}

%% file: preamble.tex
\usepackage[nonatbib,final, main]{neurips_2026}

\usepackage[utf8]{inputenc} 
\usepackage[T1]{fontenc}    
\usepackage{url}            
\usepackage{booktabs}       
\usepackage{amsfonts}       
\usepackage{nicefrac}       
\usepackage{microtype}      
\usepackage{xcolor}         

\usepackage[accsupp]{axessibility}  

\usepackage{cite}
\usepackage{caption}
\usepackage{subcaption}
\usepackage{graphicx}
\usepackage{hyperref}
\usepackage{cleveref}

\usepackage{lipsum}
\usepackage{wrapfig}
\usepackage{array}
\usepackage{xcolor}
\usepackage{makecell}
\usepackage{enumitem}

\renewcommand{\footnotesize}{\scriptsize}
\usepackage{algorithm}
\usepackage{algpseudocode}
\floatname{algorithm}{Algorithm}
\algrenewcommand\algorithmicrequire{\textbf{Require:}}
\algrenewcommand\algorithmicensure{\textbf{Ensure:}}

\usepackage{orcidlink}
\usepackage[dvipsnames]{xcolor}
\usepackage{booktabs}

\usepackage{pifont} 
\newcommand{\cmark}{\ding{51}}

\usepackage{colortbl}
\usepackage{multirow}
\usepackage{multicol}
\usepackage{listings} 
\usepackage{titletoc} 
\usepackage{fancyvrb}
\usepackage{tcolorbox} 
\usepackage[accsupp]{axessibility}  
\usepackage[subtle]{savetrees} 

\newcommand{\subpara}[1]{%
  \vspace{-1pt}%
  \noindent\textbf{#1}%
}

\definecolor{lightgray}{rgb}{0.83, 0.83, 0.83}
\definecolor{Gray}{gray}{0.6}
\definecolor{aliceblue}{rgb}{0.94, 0.97, 1.0}
\definecolor{mistyrose}{rgb}{1.0, 0.89, 0.88}
\definecolor{backcolour}{rgb}{0.95,0.95,0.92}

\newcommand{\appendixref}[2]{%
  \if\sepappendix1%
    #1
  \else%
    #2
  \fi%
}

\usepackage{cleveref}

\crefname{equation}{Eq.}{Eqs.}
\Crefname{equation}{Equation}{Equations}

\crefname{figure}{Fig.}{Figs.}
\Crefname{figure}{Figure}{Figures}

\crefname{table}{Tab.}{Tabs.}
\Crefname{table}{Table}{Tables}

\crefname{section}{Sec.}{Secs.}
\Crefname{section}{Section}{Sections}

\crefname{algorithm}{Alg.}{Algs.}
\Crefname{algorithm}{Algorithm}{Algorithms}

%% file: sections/1_intro.tex
\section{Introduction}
\label{sec:intro}
Large language models (LLMs)~\cite{achiam2023gpt, liu2024deepseek, brown2020language, bai2023qwen, comanici2025gemini, yang2025qwen3} and their multimodal extensions have demonstrated transformative performance across diverse tasks, driven by the scaling of transformer architectures~\cite{kaplan2020scaling, henighan2020scaling, zhai2022scaling, chung2024scaling, bai2025qwen3, bai2025qwen25vltechnicalreport}. Among these, Omnimodal Large Language Models (Omni-LLMs)~\cite{xu2025qwen25omnitechnicalreport, tang2025video, xu2025qwen3, team2026qwen3, hurst2024gpt}, which jointly process video, audio, and text, present a particularly challenging setting, as modality tokens from video frames and audio segments constitute the vast majority of the input sequence. At the core of these models lies the attention mechanism~\cite{vaswani2017attention}, which allows models to dynamically prioritise relevant information; yet how Omni-LLMs handle the large volume of modality tokens that dominate their inputs remains an open question.


Among the diverse behaviours observed in the attention mechanism~\cite{voita2019analyzing, olsson2022context, darcet2023vision, chefer2021transformer}, the \textit{attention sink} phenomenon has received increasing attention~\cite{xiao2023efficient}: models tend to allocate a disproportionate share of attention to semantically uninformative tokens, such as special or punctuation tokens in LLMs (\textit{e.g.,} \texttt{`<BOS>'}, `\texttt{.}', `\texttt{\textbackslash n}')~\cite{yu2024unveiling, sun2024massive, fu2025attnnotalways} and background or weakly relevant patches in vision models~\cite{kang2025see, luo2025sink, jiang2025vision, darcet2023vision}. Prior work is divided: some argue these tokens merely stabilise attention distributions and can be pruned or gated for efficiency~\cite{qiu2025gated, fu2025attnnotalways, sandoval2025identifying, guo2024active}, while others show they carry structured representations beneficial to downstream tasks~\cite{darcet2023vision, jiang2025vision, luo2025sink, sun2024massive}. In this work, we revisit these two views in the context of Omni-LLMs with complex multimodal inputs.


In this work, we present the first analysis of attention sinks 
in Omni-LLMs. While prior work in language models has largely focused on 
attention scores and token norms to characterise sink 
behaviour, we shift the focus to the representations of 
sink tokens themselves and their role during decoding. 
To this end, we formulate three fundamental questions:

\begin{itemize}[label={}, leftmargin=0pt, topsep=2pt, itemsep=1pt, parsep=0pt, before=\vspace{-4pt}, after=\vspace{-4pt}]
    \item \textbf{Q1.} Which criteria are appropriate for 
    identifying sink tokens in Omni-LLMs settings?
    \item \textbf{Q2.} Whether attention sinks solely serve 
    as redundant heads?
    \item \textbf{Q3.} Whether sink token representations 
    act as a global signal across token outputs?
\end{itemize}



For \textbf{Q1}, we examine how sink tokens should be identified in Omni-LLMs by comparing criteria proposed for LLMs~\cite{sun2024massive} and vision-language models (VLMs)~\cite{kang2025see}. We find that the VLM-based criterion classifies an excessively large fraction of tokens as sinks, including those corresponding to semantically meaningful visual objects. In contrast, the LLM-based criterion identifies only a small number of sink tokens and assigns sink behaviour to structurally defined elements such as the initial token, yielding a more stable and interpretable definition.
Building on this criterion, we address \textbf{Q2} and \textbf{Q3}. For \textbf{Q2}, we analyse the functional role of attention heads that attend heavily to sink tokens. While prior work suggests that pruning such heads preserves model performance, implying redundancy~\cite{fu2025attnnotalways, qiu2025gated, sandoval2025identifying, guo2024active, gu2024attention}, our head-by-head ablation experiments reveal no consistent trend: pruning a sink head may either improve or degrade performance depending on the case, indicating that sink heads cannot be characterised solely as inactive components.
For \textbf{Q3}, we revisit the attention output decomposition in Omni-LLMs. Since most tokens attend strongly to the sink, its value vector acts as a shared bias term added to the attention output across all tokens~\cite{sun2024massive}, meaning the direction encoded in the sink value shapes the decoding process throughout the network. 

Based on these analyses, we propose \textbf{\textit{OutRo}}, a novel inference-time decoding method for Omni-LLMs that leverages sink representations to guide token interactions during decoding. OutRo operates through two complementary mechanisms. 
First, an adaptive geometric transformation rotates the 
outputs of attention heads at non-sink positions toward 
the sink value direction, explicitly aligning token 
representations with the shared bias without requiring 
additional attention to the sink.
Second, it strengthens this global signal by selectively relaxing the causal mask on sink tokens, allowing their representations to better aggregate context from other tokens. Importantly, OutRo operates at the transformer layer level without requiring explicit access to attention maps, avoiding the memory and computational overhead of prior approaches that rely on attention manipulation~\cite{yu2024unveiling, zhang-etal-2025-shallow, huo2024self, kang2025see, wang2025gradientguided} or multiple forward passes~\cite{wang2024mitigating, kim2024code, chuang2023dola, jung2025avcd, tong2025layercontrastive}. We evaluate OutRo on seven video QA benchmarks, where it consistently improves performance with only modest decoding overhead, demonstrating that exploiting sink representations is a practical, training-free strategy for improving Omni-LLMs.

%% file: sections/5_relatedworks.tex
\section{Related Works}
\label{sec:related_works}

\subpara{Attention sinks in transformer models} refer to a phenomenon where a small set of tokens attract disproportionate attention mass despite carrying limited semantic content~\cite{xiao2023efficient, gu2024attention, queipo2025attention, zuhri2025softpick}, commonly attributed to the non-negative normalisation imposed by softmax~\cite{xiao2023efficient, sun2024massive, gu2024attention}. Two contrasting views have emerged on their role.
The first holds that sink tokens serve a purely structural role. In LLMs, since sink value representations have near-zero norms, heads attending to them produce outputs of small magnitude and are considered inactive~\cite{fu2025attnnotalways, guo2024active, sandoval2025identifying}, motivating pruning and gating strategies that remove sink-dominated heads~\cite{qiu2025gated, sok2026garbage, sandoval2025identifying}. In VLMs, visual sink tokens are similarly treated as uninformative patches, and redistributing their attention to more relevant regions has been shown to improve performance~\cite{kang2025see, jiao2025don, tang2025duallevelattention}.
The second view suggests that sink tokens carry meaningful representations. In ViTs, high-norm sink tokens aggregate global information when full attention is available~\cite{darcet2023vision, jiang2025vision}. In vision-language models, whether a visual sink originates from the ViT or emerges within the LLM determines what it encodes, with ViT-propagated sinks capturing coarse-grained global context~\cite{luo2025sink}. In LLMs, sink tokens act as implicit bias terms in the attention output, injecting a shared direction across all token representations~\cite{sun2024massive}. 

\subpara{Inference-time methods for multimodal LLMs (MLLMs)} focus on enhancing multimodal reasoning and robustness in a training-free manner. One array of work builds upon contrastive decoding paradigms, which improve predictions by contrasting outputs obtained under different conditions, such as perturbed inputs, alternative decoding paths, or representations from different model layers~\cite{kim2024code, jung2025avcd, leng2024mitigating, chuang2023dola, zhang2025activelayercontrastive, tong2025layercontrastive}. These methods suppress hallucinations and spurious correlations by comparing biased and unbiased predictions. Another range of work treats attention sinks as explicit intervention targets at inference time. By analysing attention maps, these methods identify sink tokens and suppress their influence by redistributing attention mass towards more informative tokens~\cite{yu2024unveiling, kang2025see, jiao2025don, tang2025duallevelattention, jung2026probing}. However, these strategies incur overhead: contrastive decoding requires multiple forward passes, while sink-aware methods rely on attention map access or activation manipulation, thus not tailored to optimised attention implementations~\cite{dao2022flashattentionfastmemoryefficientexact, dao2023flashattention2fasterattentionbetter, shah2024flashattention3fastaccurateattention}. Our work addresses this gap by proposing a modality-agnostic, 
training-free intervention for Omni-LLMs that operates at 
the level of attention head outputs, without relying on 
attention maps or repeated inference.

%% file: sections/2_attentionsink.tex
\section{On the Nature of Attention Sink}
\label{sec:AttentionSink}

Unless otherwise specified, all experiments in this 
section are conducted on Qwen2.5-Omni~\cite{xu2025qwen25omnitechnicalreport}.

\subpara{Preliminaries of Omni-LLMs.}
Omni-LLMs process a video input $\mathcal{V}$, an audio stream $\mathcal{A}$, and a textual instruction $\mathcal{L}$. Modality encoders transform these into sequences of embeddings: $\mathcal{V}$ and $\mathcal{A}$ are projected into $M$ modality tokens $\mathbf{X}_m \in \mathbb{R}^{M \times D}$ encapsulating visual and auditory features, which are concatenated with $L$ textual tokens $\mathbf{X}_l \in \mathbb{R}^{L \times D}$ from a language model tokenizer to form the input:
\begin{equation}
    \mathbf{X} = [\mathbf{X}_m; \mathbf{X}_l] \in 
    \mathbb{R}^{N \times D}, \quad N = M + L,
\end{equation}
where $D$ denotes the hidden dimension of the transformer.

\subpara{Multi-head attention.}
Each transformer layer in Omni-LLMs contains a multi-head attention (MHA) module. For each head $h \in \{1, \dots, H\}$, the input $\mathbf{X}$ is projected into queries $\mathbf{Q}_h$, keys $\mathbf{K}_h$, and values $\mathbf{V}_h$ via weight matrices $\mathbf{W}_Q, \mathbf{W}_K, \mathbf{W}_V \in \mathbb{R}^{D \times D_h}$. A causal mask $\mathcal{M}$ enforces autoregressive generation by setting $\mathcal{M}_{ij} = -\infty$ for $j > i$. The attention distribution for head $h$ is then defined as:
\begin{equation}
    A_{h,i,:} = \text{Softmax}\left(
        \frac{\mathbf{Q}_{h,i}\mathbf{K}_h^\top}{\sqrt{D_h}} 
        + \mathcal{M}_{i,:}
    \right),
    \label{eq:2}
\end{equation}
and the output representation for each head is computed as:
\begin{equation}
    \mathbf{O}_h = \mathbf{A}_h \mathbf{V}_h,
    \; \text{where} \;
    \mathbf{O}_{h,i} = \sum_{j=1}^{i} A_{h,i,j}\mathbf{v}_{h,j}.
\end{equation}
The final representation of the layer is obtained by 
concatenating all head outputs and applying a linear 
projection $\mathbf{W}_O$:
$\mathbf{O}_{final} = [\mathbf{O}_1; \mathbf{O}_2; 
    \dots; \mathbf{O}_H] \mathbf{W}_O.$



\begin{itemize}[label={}, leftmargin=0pt, topsep=4pt, itemsep=4pt, after=\vspace{-8pt}]
\item \textbf{Q1. Which criteria are appropriate for identifying sink tokens in Omni-LLMs settings?}
\end{itemize}
\label{subsec:q1}
The criterion originates from analyses on LLMs, where sink phenomena were characterized through \textit{massive activations}~\cite{sun2024massive} in hidden states across layers. Specifically, given the set of token hidden states in a layer $\mathbf{X}=\{\mathbf{x}_1,\dots,\mathbf{x}_N\}$ with 
$\mathbf{x}_i\in\mathbb{R}^{1\times D}$, an activation scalar $z = \mathbf{x}_i[d]$ for some $i \in \{1,\dots,N\}$ and $d \in \{1,\dots,D\}$ is regarded as massive if
\begin{equation}
    \Phi_{\text{LLM}}(\mathbf{x}_i)
    = \max_{d\in\{1,\dots,D\}} |\mathbf{x}_i[d]|
    > \max\!\big(100,\; 1000 \times 
    \text{median}_{\mathbf{z}\in\mathbf{X}}
    (|\mathbf{z}|)).
    \label{eq:5}
\end{equation}

\vspace{-8pt}
More recently, this notion was extended to \textit{visual attention sink} by examining sink dimensions inherited from LLMs~\cite{kang2025see}. Let $D_{\text{sink}}$ denote a fixed set of dimensions known to exhibit sink behaviour in the underlying LLM. For a hidden state $\mathbf{x}_i\in\mathbb{R}^{1\times D}$, the sink dimension value was defined as:
\begin{equation}
    \Phi_{\text{VLM}}(\mathbf{x}_i) =
    \max_{d_s\in D_{\text{sink}}}
    \left|
        \frac{\mathbf{x}_i[d_s]}
        {\sqrt{\frac{1}{D}\sum_{d=1}^{D}\mathbf{x}_i[d]^2}}
    \right|
    \ge \tau = 20.
    \label{eq:6}
\end{equation}

\begin{figure}[t]
\centering
\captionsetup[subfigure]{aboveskip=0pt}
\scalebox{0.95}{%
\begin{minipage}{\linewidth}
\centering
\begin{subfigure}{0.46\linewidth}
    \centering
    \includegraphics[width=\linewidth]{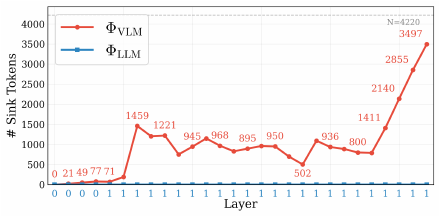}
    \vspace{-4mm}
    \caption{}
    \label{fig:fig1a}
\end{subfigure}
\hfill
\begin{subfigure}{0.50\linewidth}
    \centering
    \includegraphics[width=\linewidth]{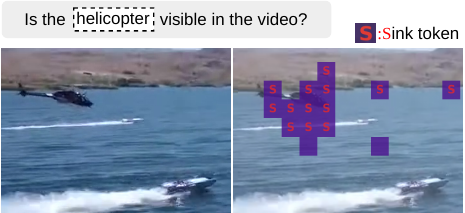}
    \vspace{-4mm}
    \caption{}
    \label{fig:fig1b}
\end{subfigure}
\end{minipage}%
}
\vspace{-5pt}
\caption{\textbf{$\Phi_{\text{VLM}}$ incorrectly identifies semantic tokens as sinks.}
    \textbf{(a)} $\Phi_{\text{VLM}}$ rapidly over-identifies sink tokens in deeper layers, unlike the sparse behaviour of $\Phi_{\text{LLM}}$.
    \textbf{(b)} For the query ``Is the \textit{helicopter} visible?'', highly attended object tokens are still identified as sinks under $\Phi_{\text{VLM}}$.
}
\vspace{-3mm}
\label{fig:fig1}
\end{figure}

\input{tables/sink_outlier_dims}
\subpara{Observation.}
\cref{fig:fig1a} reveals a striking degeneracy when the visual attention sink criterion~$\Phi_{\text{VLM}}$ is applied to Omni-LLMs.
As depth increases, the majority of tokens are classified as sinks, eventually encompassing even semantically meaningful ones.
This suggests that $\Phi_{\text{VLM}}$ fails to discriminatively characterise sink tokens in Omni-LLM settings: rather than isolating a sparse subset of anomalous tokens, the criterion collapses, marking the bulk of representations as sinks
(See~\cref{supp:q1_qualitative} of our appendix for more examples).

To investigate this discrepancy, \cref{tab:sink_outlier_dims} compares dimensions identified via the outlier feature criterion~\cite{dettmers2022gpt3, sun2024massive} with the sink dimensions $D_{\text{sink}}$ in \cref{eq:6}, defined over the $3584$-dimensional hidden space. An activation is treated as an outlier if its magnitude exceeds 6.0 in more than 25\% of layers and 6\% of tokens across at least 90 out of 100 evaluation sequences. All dimensions in $D_{\text{sink}}$ are also identified as outliers, suggesting that sink dimensions largely correspond to norm-amplifying patterns. In contrast, $\Phi_{\text{LLM}}$ from \cref{eq:5} yields a sparse and stable set of sinks, typically one dominant token per layer, shown to be critical for model performance~\cite{xiao2023efficient, yu2024unveiling}, indicating it provides a more robust criterion for Omni-LLMs. Detailed results are provided in~\cref{supp:sink_stats,supp:sink_empirical} of our appendix.

Overall, these findings demonstrate that $\Phi_{\text{VLM}}$ does not generalise reliably to Omni-LLM settings. To obtain a more general identification of sink behaviour, we adopt $\Phi_{\text{LLM}}$ for subsequent analyses.

\begin{itemize}[label={}, leftmargin=0pt, topsep=4pt, itemsep=4pt, after=\vspace{-8pt}]
\item \textbf{Q2. Whether attention sinks solely serve as redundant heads?}
\end{itemize}
\label{subsec:q2}

A prevailing interpretation of attention sinks is that they are functionally negligible due to the near-zero norms of their value representations~\cite{qiu2025gated, fu2025attnnotalways, guo2024active}. Under this view, even if a head assigns substantial attention to sink tokens, the resulting head output remains relatively small, and the head is therefore considered effectively suppressed. This perspective has motivated pruning strategies that remove sink-dominated heads under the assumption that they are redundant.

\begin{figure}[t]
\centering
\captionsetup[subfigure]{aboveskip=3pt}
\scalebox{0.95}{%
\begin{minipage}{\linewidth}
\centering
\begin{subfigure}{0.33\linewidth}
    \centering
    \includegraphics[width=\linewidth]{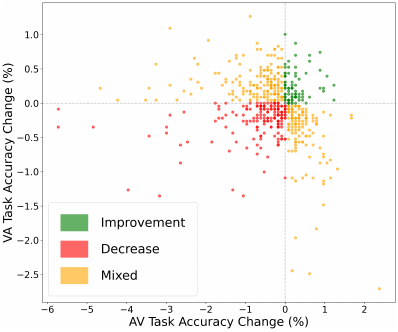}
    \vspace{-15pt}
    \caption{}
    \label{fig:fig6a}
\end{subfigure}
\hfill
\begin{subfigure}{0.64\linewidth}
    \centering
    \includegraphics[width=\linewidth]{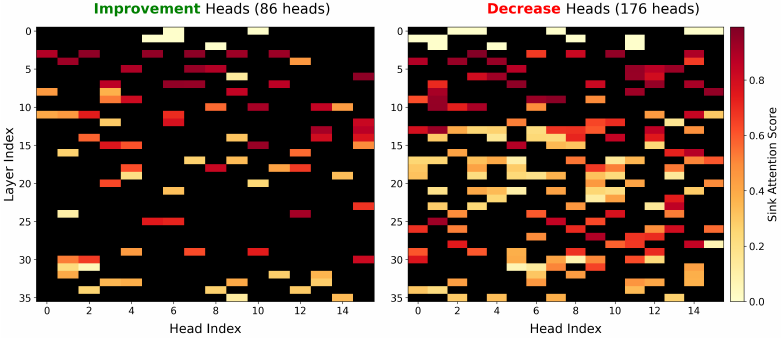}
    \vspace{-15pt}
    \caption{}
    \label{fig:fig6b}
\end{subfigure}
\end{minipage}%
}
\vspace{-2mm}
\caption{\textbf{High sink attention does not imply head redundancy.}
\textbf{(a)} Performance change after removing individual heads, distinguishing \textcolor{green!60!black}{improvement} and \textcolor{red!70!black}{decrease} heads.
\textbf{(b)} High sink scores appear in both \textcolor{green!60!black}{improvement} and \textcolor{red!70!black}{decrease} heads, indicating sink-heavy heads are not always redundant.}
\label{fig:fig6}
\vspace{-5mm}
\end{figure}

\input{tables/head_prune_ablation}
\subpara{Observation.}
Prior work has explored pruning strategies based on sink attention scores~\cite{sandoval2025identifying, sok2026garbage}, under the assumption that heads attending heavily to sink tokens are largely redundant. However, our head-by-head ablation experiments on Omni-LLMs reveal a different picture. We evaluate on AVHBench~\cite{sung2024avhbench}, which comprises an audio-driven video hallucination subset ($A \to V$) and a video-driven audio hallucination subset ($V \to A$). As shown in \cref{fig:fig6}, removing certain sink heads leads to marked performance improvements, while removing others causes clear degradation, with no consistent trend across cases. \cref{tab:head_change} further confirms this: heads with comparably high sink attention scores appear in both groups yet yield opposite performance impacts. These results demonstrate that high sink attention alone is insufficient to determine head importance, suggesting that sink value representations encode functional information beyond mere attention absorption.


\begin{figure}[t]
\centering
\scalebox{0.88}{%
\begin{minipage}{\linewidth}
\centering
\begin{subfigure}{0.48\linewidth}
\centering
\includegraphics[width=\linewidth]{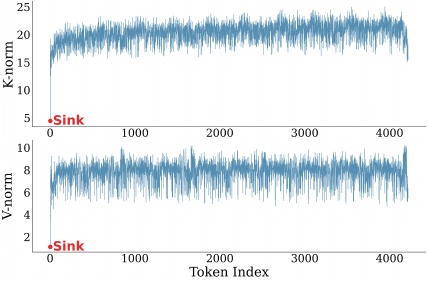}
\vspace{-4mm}
\vspace{-4pt}
\caption{}
\label{fig:fig2a}
\end{subfigure}
\begin{subfigure}{0.48\linewidth}
\centering
\includegraphics[width=\linewidth]{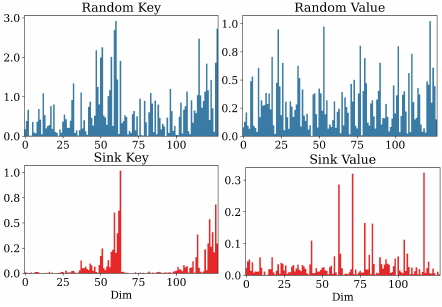}
\vspace{-4mm}
\vspace{-4pt}
\caption{}
\label{fig:fig2b}
\end{subfigure}
\begin{subfigure}{0.52\linewidth}
\centering
\includegraphics[width=\linewidth]{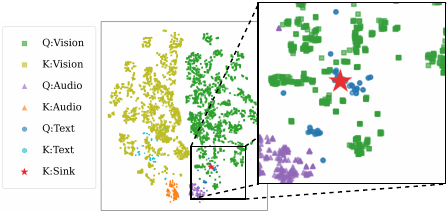}
\vspace{-4mm}
\vspace{-4pt}
\caption{}
\label{fig:fig2c}
\end{subfigure}
\begin{subfigure}{0.44\linewidth}
\centering
\includegraphics[width=\linewidth]{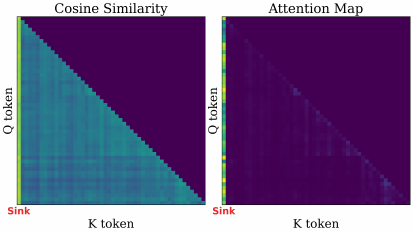}
\vspace{-4mm}
\vspace{-4pt}
\caption{}
\label{fig:fig2d}
\end{subfigure}
\end{minipage}%
}
\vspace{-2mm}
\caption{
\textbf{Geometric analysis of sink token representations.}
\textbf{(a)} Sink keys and values have substantially smaller norms than other tokens, yet \textbf{(b)} their activations are not uniformly small across dimensions.
\textbf{(c)} In t-SNE, sink keys cluster near the query distribution, and \textbf{(d)} their cosine similarity with queries remains high, yielding large attention scores.
}
\label{fig:fig2}
\vspace{-4mm}
\end{figure}

\begin{itemize}[label={}, leftmargin=0pt, topsep=4pt, itemsep=4pt, after=\vspace{-8pt}]
\item \textbf{Q3. Whether sink token representations act as a global signal across token outputs?}
\end{itemize}
\label{subsec:q3}

While \textbf{Q2} suggests that small-magnitude sink value representations do not imply redundancy, we make this explicit by analysing sink token representations in the feature space. Noting that sink key representations similarly exhibit near-zero norms, we first examine whether such representations can nonetheless carry meaningful structure, before turning to their value counterparts.

\begin{wrapfigure}{r}{0.48\linewidth}
\centering
\vspace{-10pt}
\includegraphics[width=\linewidth]{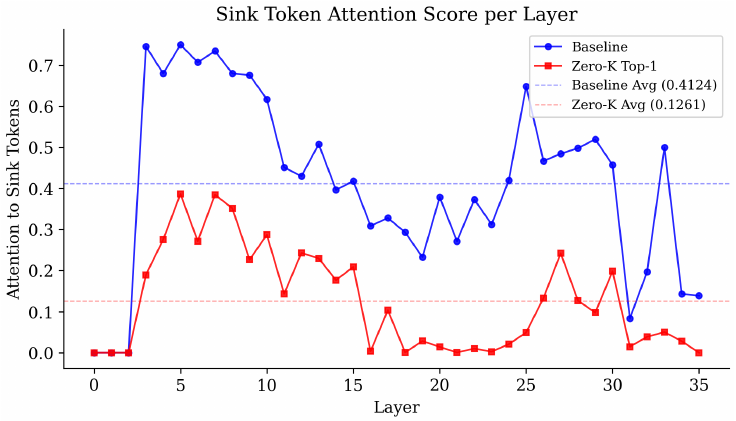}
\vspace{-15pt}
\caption{\textbf{Sink token attention score.} Compared to the baseline, zeroing the top-1 dimension in the sink key representations substantially reduces attention to sink tokens across layers.}
\label{fig:fig4}
\vspace{-15pt}
\end{wrapfigure}
\begin{figure}[t]
    \centering
    \includegraphics[width=0.9\linewidth]{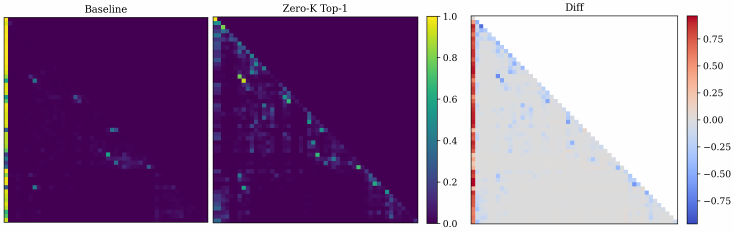}
    \vspace{-3mm}
    \caption{\textbf{Zeroing the top-1 sink key dimension (Zero-K).} 
    Attention maps before (Baseline) and after (Zero-K Top-1) 
    the intervention show that suppressing the dominant sink 
    key dimension effectively redirects attention away from 
    sink positions, as confirmed by the difference map.}
\label{fig:fig3}
\vspace{-5mm}
\end{figure}

\subpara{Observation.}
As shown in \cref{fig:fig2a}, sink key and value representations exhibit smaller norms than other tokens, yet remain structured rather than uniformly suppressed (\cref{fig:fig2b}). Moreover, sink keys are geometrically aligned with query representations, showing high cosine similarity and attention scores despite their reduced magnitudes (\cref{fig:fig2c,fig:fig2d}). These observations suggest that low norm does not imply irrelevance.
To further validate this, we examine the internal 
structure of sink key activations, which are not 
uniformly suppressed (\cref{fig:fig2b}).
We hypothesise that these dimensions drive sink behaviour, such that zeroing them should disrupt the characteristic attention pattern. We test this by zeroing out the top-magnitude dimensions in sink key representations, which we term Zero-K. As shown in \cref{fig:fig3,fig:fig4}, zeroing only the top-1 dimension substantially eliminates the attention sink pattern, with attention to sink positions becoming consistently suppressed across layers.

\begin{figure}[t]
\centering
\scalebox{0.93}{%
\begin{minipage}{\linewidth}
\centering
\begin{subfigure}{0.32\linewidth}
    \centering
    \includegraphics[width=\linewidth]{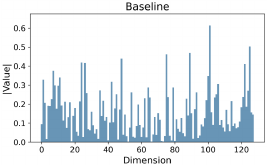}
    \vspace{-5mm}
    \caption{}
    \label{fig:fig7a}
\end{subfigure}
\hfill
\begin{subfigure}{0.32\linewidth}
    \centering
    \includegraphics[width=\linewidth]{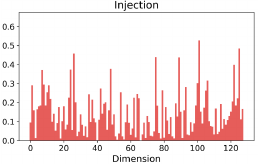}
    \vspace{-5mm}
    \caption{}
    \label{fig:fig7b}
\end{subfigure}
\hfill
\begin{subfigure}{0.32\linewidth}
    \centering
    \includegraphics[width=\linewidth]{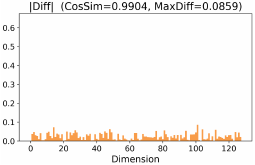}
    \vspace{-5mm}
    \caption{}
    \label{fig:fig7c}
\end{subfigure}
\end{minipage}%
}
\vspace{-5pt}
\caption{\textbf{Sink token head outputs before and after injection.}
The outputs remain nearly identical, with only minimal differences observed at the last layer.}
\label{fig:fig7}
\vspace{-2mm}
\end{figure}

Notably, although Zero-K effectively removes the sink attention pattern, it does not improve performance; as reported in \cref{tab:zero-k}, removing dominant sink dimensions consistently leads to performance degradation. These results demonstrate that sink key representations, despite their low magnitudes, carry meaningful information that shapes the attention mechanism.

\input{tables/q3_tables}

\subpara{Hypothesis.}
These findings raise the question of what information is 
carried by sink value representations. Following the 
attention output decomposition of~\cite{sun2024massive}, 
since most tokens attend strongly to the sink, the 
attention output at each token can be decomposed as:
\begin{equation}
    \mathbf{O}_{h,i} = \sum_{j \in \mathcal{S}} A_{h,i,j} 
    \mathbf{v}_{h,j} + \sum_{j \notin \mathcal{S}} 
    A_{h,i,j} \mathbf{v}_{h,j},
\end{equation}
where the first term, the value update from sink positions, acts as a shared bias term added uniformly across all token outputs. We hypothesise that this bias encodes a structured direction that sink-attending heads align with and propagate across layers, and examine this from two perspectives: the role of sink positions themselves, and the effect of aligning non-sink positions with the sink direction.
\begin{wrapfigure}{r}{0.48\linewidth}
\centering
\vspace{-10pt}
\includegraphics[width=\linewidth]{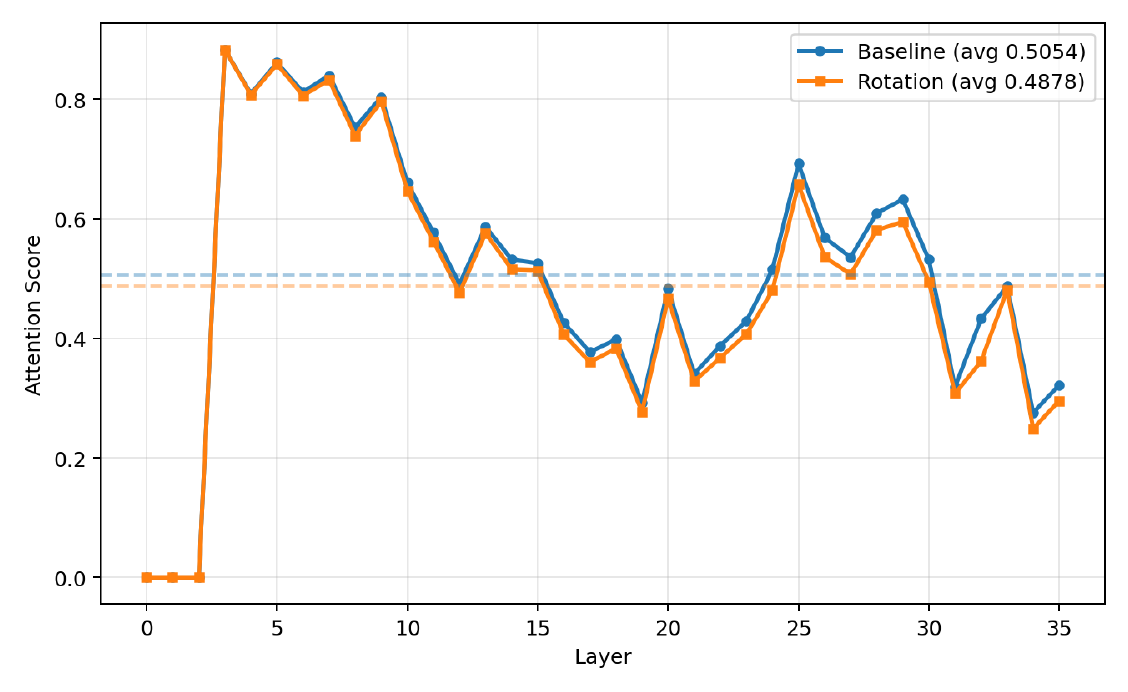}
\vspace{-20pt}
\caption{\textbf{Attention to sink tokens before and after rotation.} 
Rotation consistently reduces the total attention 
allocated to sink tokens.}
\label{fig:attn_shift}
\vspace{-15pt}
\end{wrapfigure}
\textbf{For sink positions}, we examine whether explicitly 
enriching sink representations sharpens this bias 
direction by relaxing causal masking from layers where 
sink behaviour first emerges through to the second-to-last 
layer, allowing contextual representations to flow into 
sink positions. As shown in 
\cref{fig:fig7a,fig:fig7b,fig:fig7c}, the resulting 
representations are nearly identical to the original, 
indicating that the bias direction is already stably 
encoded prior to injection. Nevertheless, 
\cref{tab:crossattn} shows that enriching sink 
representations yields consistent performance gains, 
suggesting that even small refinements to the bias 
direction are beneficial.
\textbf{For non-sink positions}, we rotate head outputs 
toward the sink value direction (i.e., projecting each 
head output onto the sink value vector, adding the 
projected component, and rescaling to preserve the 
original magnitude) and observe consistent performance 
improvements (\cref{tab:head_rotation}). As shown in 
\cref{fig:attn_shift}, rotation reduces attention 
allocated to sink tokens, with the freed attention 
redistributing toward modality tokens. Notably, this 
redistribution is not uniform: attention shifts 
preferentially toward the modality that the model 
originally attended to less (see~\cref{supp:rotate_attn} of our appendix for details),
suggesting that by explicitly providing the sink bias to 
non-sink positions, the model no longer needs to attend 
to the sink to obtain it, and the released attention 
capacity is used to better balance across modalities.

%% file: tables/sink_outlier_dims.tex


\begin{wraptable}{r}{0.35\linewidth}
\vspace{-4mm}
\centering
\footnotesize
\setlength{\tabcolsep}{12pt}
\renewcommand{\arraystretch}{1.0}
\caption{\textbf{Sink/outlier dimensions.}
Sink dimensions strongly overlap with outlier feature patterns.}
\vspace{-3mm}
\label{tab:sink_outlier_dims}
\resizebox{\linewidth}{!}{
\begin{tabular}{@{}ll@{}}
\toprule
 & \textbf{Dims} \\
\midrule
\textbf{Sink}    & 458, 2570, \ldots \\
\midrule
\textbf{Outlier} & \textbf{458}, \textbf{2570}, 3206, 3281, 1427, \ldots \\
\bottomrule
\end{tabular}
}
\vspace{-5mm}
\end{wraptable}

%% file: tables/head_prune_ablation.tex
\begin{wraptable}{r}{0.48\linewidth}
\vspace{-13pt}
\centering
\scriptsize
\setlength{\tabcolsep}{8pt}
\renewcommand{\arraystretch}{1.05}
\caption{\textbf{Head ablation performance.} Heads with similar sink scores can yield opposite performance changes.}
\vspace{-2mm}
\label{tab:head_change}
\begin{tabular}{@{}l c c c c@{}}
\toprule
\textbf{Head} & Sink score & $A \to V$ & $V \to A$ & Overall \\
\midrule
L15H3  & 0.7553 & \textcolor{green!60!black}{\texttt{+}0.88\%} & \textcolor{green!60!black}{\texttt{+}0.48\%} & \textcolor{green!60!black}{\texttt{+}1.36\%} \\
L18H12 & 0.6633 & \textcolor{green!60!black}{\texttt{+}0.26\%} & \textcolor{green!60!black}{\texttt{+}0.87\%} & \textcolor{green!60!black}{\texttt{+}1.14\%} \\
L11H6  & 0.7003 & \textcolor{green!60!black}{\texttt{+}0.62\%} & \textcolor{green!60!black}{\texttt{+}0.09\%} & \textcolor{green!60!black}{\texttt{+}0.70\%} \\
\midrule
L24H10 & 0.7755 & \textcolor{red!70!black}{-3.43\%} & \textcolor{red!70!black}{-0.48\%} & \textcolor{red!70!black}{-3.91\%} \\
L16H14 & 0.8773 & \textcolor{red!70!black}{-2.55\%} & \textcolor{red!70!black}{-0.13\%} & \textcolor{red!70!black}{-2.68\%} \\
L14H12 & 0.7497 & \textcolor{red!70!black}{-1.32\%} & \textcolor{red!70!black}{-0.44\%} & \textcolor{red!70!black}{-1.76\%} \\
\toprule
\end{tabular}
\vspace{-20pt}
\end{wraptable}

%% file: tables/q3_tables.tex
\begin{figure}[t]
\begin{minipage}[t]{0.48\linewidth}
\centering
\scriptsize
\setlength{\tabcolsep}{6pt}
\captionof{table}{\textbf{Zero-K performance.} 
Removing dominant sink dimensions degrades performance on AVUT~\cite{yang2025audio} and DailyOmni~\cite{zhou2025daily} across both Qwen2.5-Omni and SALMONN2+, suggesting that sink key representations carry functional information that shapes the attention mechanism.}
\label{tab:zero-k}
\vspace{-2mm}
\resizebox{\linewidth}{!}{
\begin{tabular}{l cc cc}
\toprule
\multirow{2.5}{*}{\textbf{Acc.$\uparrow$}}
& \multicolumn{2}{c}{Qwen2.5-Omni~\cite{xu2025qwen25omnitechnicalreport}} 
& \multicolumn{2}{c}{SALMONN2\texttt{+}\cite{tang2025video}} \\
\cmidrule(lr){2-3} \cmidrule(lr){4-5}
& AVUT & DailyOmni & AVUT & DailyOmni \\
\midrule
\rowcolor{gray!8}
Baseline & 65.79 & 54.14 & 56.73 & 51.71 \\
Top-1    & 61.31 & 48.62 & 51.07 & 37.76 \\
Top-5    & 59.94 & 47.28 & 40.16 & 25.98 \\
Top-10   & 58.87 & 45.45 & 37.82 & 25.90 \\
\bottomrule
\end{tabular}}
\end{minipage}
\hfill
\begin{minipage}[t]{0.48\linewidth}
\centering
\scriptsize
\setlength{\tabcolsep}{8pt}
\renewcommand{\arraystretch}{1.0}
\captionof{table}{\textbf{Information injection} into sink tokens leads to consistent gains.}
\label{tab:crossattn}
\vspace{-2.5mm}
\resizebox{\linewidth}{!}{
\begin{tabular}{@{}l cc cc@{}}
\toprule
\multirow{2.5}{*}{\textbf{Acc.$\uparrow$}}
& \multicolumn{2}{c}{Qwen2.5-Omni} 
& \multicolumn{2}{c}{SALMONN2\texttt{+}} \\
\cmidrule(lr){2-3} \cmidrule(lr){4-5}
& AVH & DailyOmni & AVH & DailyOmni \\
\midrule
Baseline & 70.91 & 55.38 & 62.07 & 51.71 \\
Injection & \textbf{71.08} & \textbf{55.47} & \textbf{62.20} & \textbf{51.96} \\
\bottomrule
\end{tabular}
}
\vspace{-1mm}
\setlength{\tabcolsep}{10pt}
\renewcommand{\arraystretch}{1.0}
\captionof{table}{\textbf{Head output rotation} toward the sink value direction improves performance.}
\label{tab:head_rotation}
\vspace{-1mm}
\resizebox{\linewidth}{!}{
\begin{tabular}{l c c c}
\toprule
\textbf{AVHBench} & $A \to V$ & $V \to A$ & Matching \\
\midrule
Baseline & 80.11 & 75.41 & 59.87 \\
Rotation & \textbf{80.81} & \textbf{75.72} & \textbf{60.50} \\
\bottomrule
\end{tabular}}
\end{minipage}
\vspace{-5mm}
\end{figure}

%% file: sections/3_method.tex
\section{Method}
\label{sec:Method}

Based on our analysis, we introduce \textbf{\textit{OutRo}}, a unified inference-time modulation method that combines (i) ReLU-tanh gating to align non-sink representations with the sink bias direction and (ii) sink information enhancement via one-time mask relaxation. The method operates without parameter updates and is implemented via lightweight forward interventions on attention head outputs.

\subpara{Sink value direction.}
At each layer~$\ell$, sink token indices $\mathcal{S}^{(\ell)}$ 
are identified at inference time using $\Phi_{\mathrm{LLM}}$ 
(\cref{eq:5}). For each head~$h$, the value representations 
at these positions are averaged as:
\begin{equation}
  \bar{\mathbf{v}}_{h, \mathcal{S}}^{(\ell)}
  \;=\;
  \frac{1}{|\mathcal{S}^{(\ell)}|}
  \sum_{s \in \mathcal{S}^{(\ell)}} 
  \mathbf{v}_{h,s}^{(\ell)}\,.
\end{equation}

\subpara{ReLU--tanh gated head output rotation.}
Our finding that aligning head outputs with the sink 
direction improves performance (\cref{tab:head_rotation}) 
motivates a gated rotation mechanism.
For efficiency, we adopt a directional-alignment proxy 
(see~\cref{supp:proxy_sink} of our appendix) for the sink score.
Specifically, we compute, for each head~$h$ and non-sink 
position~$ns \notin \mathcal{S}^{(\ell)}$, the directional 
alignment between the head output $\mathbf{O}_{h,ns}$ and 
the sink value direction:
\begin{equation}
  c_{h,ns}
  =
  \cos\!\Big(\mathbf{O}_{h,ns},\;
             \bar{\mathbf{v}}_{h, \mathcal{S}}^{(\ell)}\Big).
  \label{eq:proxy}
\end{equation}
To control the rotation strength, we convert this 
alignment into a soft gate:
\begin{equation}
  g_{h,ns}
  =
  \tanh\!\left(\frac{\operatorname{ReLU}(c_{h,ns})}{t}\right),
  \quad \text{where } t = 0.1.
  \label{eq:gate}
\end{equation}
We then rotate the head output by adding a gated 
projection onto sink direction:
\begin{equation}
\hat{\mathbf{O}}_{h,ns}
=
\mathbf{O}_{h,ns}
+
\gamma g_{h,ns}
\frac{\mathbf{O}_{h,ns} \cdot \bar{\mathbf{v}}_{h, \mathcal{S}}^{(\ell)}}
     {\|\bar{\mathbf{v}}_{h, \mathcal{S}}^{(\ell)}\|^{2}}
\bar{\mathbf{v}}_{h, \mathcal{S}}^{(\ell)},
\qquad
\mathbf{O}^{\mathrm{rotated}}_{h,ns}
=
\frac{\|\mathbf{O}_{h,ns}\|}
     {\|\hat{\mathbf{O}}_{h,ns}\|}
\hat{\mathbf{O}}_{h,ns},
\end{equation}
where $\gamma > 0$ controls the rotation strength.
This gating selectively and adaptively strengthens heads 
that are strongly aligned with the sink value direction.

\begin{figure}[t]
    \centering
    \includegraphics[width=0.9\linewidth]{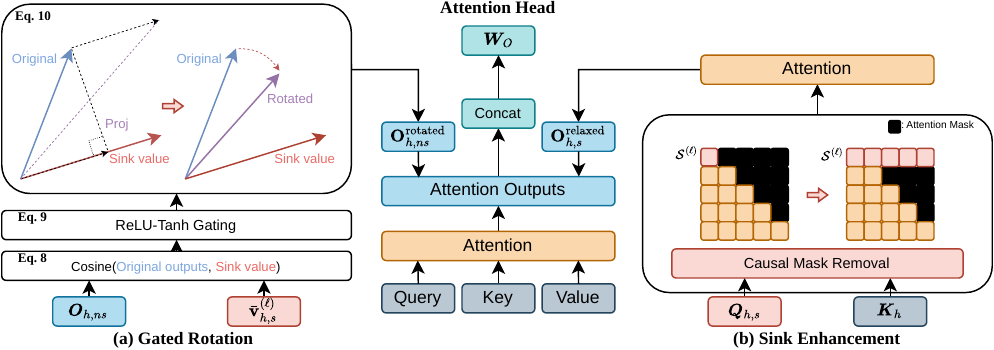}
    \vspace{-2mm}
    \caption{\textbf{Overall OutRo pipeline.}
    \textbf{(a)} Gated head output rotation. Non-sink outputs 
    are rotated toward the sink value direction via adaptive 
    gating.
    \textbf{(b)} Sink enhancement. The causal mask is removed 
    for sink positions, sharpening the shared bias direction.}
\label{fig:method}
\vspace{-3mm}
\end{figure}

\subpara{Sink information enhancement via mask relaxation.}
Our analysis in \cref{subsec:q3} showed that sink tokens 
already encode the bias direction stably, and that 
enriching this via causal mask relaxation yields 
consistent improvements (\cref{tab:crossattn}). 
We implement this by relaxing the causal mask at a 
selected layer $\ell_{\mathrm{enh}}$. At layer 
$\ell_{\mathrm{enh}}$, attention for sink positions 
$s \in \mathcal{S}^{(\ell_{\mathrm{enh}})}$ is computed 
without the causal constraint. Concretely, for sink 
positions, we remove the causal mask and compute
\begin{equation}
\mathbf{A}^{\mathrm{relaxed}}_{h,s,:}
=
\mathrm{Softmax}\!\left(
\frac{\mathbf{Q}_{h,s}\mathbf{K}_h^\top}{\sqrt{D_h}}
\right),
\end{equation}
allowing sink queries to attend to the full sequence. 
The relaxed attention is then used to compute the 
corresponding head outputs
\begin{equation}
\mathbf{O}^{\mathrm{relaxed}}_{h,s}
=
\mathbf{A}^{\mathrm{relaxed}}_{h,s,:}\mathbf{V}_h,
\end{equation}
which replace the masked outputs at sink positions.

Together, these components allow OutRo to improve model 
performance at inference time without modifying attention 
maps or requiring iterative forward passes. Importantly, 
OutRo is fully compatible with optimised attention 
implementations such as 
FlashAttention\footnote{\url{https://github.com/Dao-AILab/flash-attention}}, 
making it a practical and scalable inference-time 
strategy across diverse multimodal configurations.

%% file: sections/4_experiments.tex
\section{Experiments}
\label{sec:Experiments}

\subsection{Experimental Setup}

\subpara{Baselines.}
We evaluate on representative Omni-LLMs, including 
Qwen2.5-Omni~\cite{xu2025qwen25omnitechnicalreport} and 
video-SALMONN2\texttt{+}~\cite{tang2025video}. All models 
are evaluated using greedy decoding.

\subpara{Datasets and metrics.}
We evaluate OutRo on seven video QA benchmarks across audio-visual and visual-only settings. For audio-visual QA, we report results on OmniBench~\cite{li2024omnibench}, AVUT~\cite{yang2025audio}, AVHBench~\cite{sung2024avhbench}, and DailyOmni~\cite{zhou2025daily}, covering cross-modal consistency, hallucination detection, and audio-visual reasoning. For visual-only QA, we evaluate on VideoHolmes~\cite{cheng2025video}, VideoMME (medium)~\cite{fu2025video}, and ActivityNetQA~\cite{yu2019activitynet}, focusing on temporal reasoning and event understanding. Accuracy is reported on all benchmarks.


\subpara{Implementation details.}
For gated rotation, we search the rotation strength 
$\gamma \in [0.5, 4.0]$ and fix a single value per model: 
$\gamma = 0.5$ for video-SALMONN2\texttt{+} and 
$\gamma = 3.0$ for Qwen2.5-Omni. Rotation is applied to 
non-sink positions at all layers where sink tokens are 
present, except the final few layers. Sink information 
enhancement is applied once at approximately one-seventh 
of total depth: layer 4 for Qwen2.5-Omni$_{7\mathrm{B}}$ 
(28 layers), and layer 5 for Qwen2.5-Omni$_{3\mathrm{B}}$ 
and video-SALMONN2\texttt{+} (36 layers). All experiments 
use a single NVIDIA RTX A6000 GPU.

\subsection{Experimental Results}
\label{subsec:experimental_results}
\subpara{Results.}
\cref{tab:outro_avllm} reports results on seven video QA.
Across Qwen2.5-Omni and video-SALMONN2\texttt{+}, OutRo consistently improves performance over base models on both audio-visual and visual-only QA benchmarks. For Qwen2.5-Omni$_{7\mathrm{B}}$, OutRo achieves the largest gain on AVHBench (\texttt{+}2.18), which explicitly evaluates audio-visual hallucination.
Consistent improvements are observed on OmniBench and AVUT.
Similarly, Qwen2.5-Omni$_{3\mathrm{B}}$ benefits from gains across all audio-visual benchmarks. For video-SALMONN2\texttt{+}, OutRo improves performance across most settings, including both audio-visual and visual-only benchmarks. Importantly, these consistent gains across video benchmarks demonstrate that OutRo remains effective under complex cross-modal interactions.

\input{tables/outro_avllm}

\subsection{Further Analysis}
Here, we conduct all analysis below with the Qwen2.5-Omni model.

\input{tables/decoding_layench_compat_cd}
\subpara{Decoding efficiency.}
To assess the computational overhead of OutRo, we measure 
decoding latency in seconds per token on 100 examples from 
AVHBench (\cref{tab:latency}). OutRo increases latency 
from 0.98 to 1.09 sec$/$token, a 1.11$\times$ slowdown 
relative to baseline. In contrast, AVCD~\cite{jung2025avcd}, 
the Omni-LLM extension of contrastive decoding, is the most 
expensive, requiring up to four forward passes per step. 
VCD~\cite{leng2024mitigating} nearly doubles inference time 
with two forward passes per step, and VAR~\cite{kang2025see} 
incurs overhead from explicit attention map modification 
that disables optimised attention implementations.

\subpara{OutRo with contrastive decoding.}
\cref{tab:ablation_cd} reports results when combining 
OutRo with existing contrastive decoding (CD) methods. 
For Qwen2.5-Omni$_{3\mathrm{B}}$\footnote{Due to GPU memory constraints when using eager attention, we resize video inputs (256 $\times$ 256 ) for~\cref{tab:ablation_cd} experiments, which
results in slightly lower baseline performance compared to the main results in~\cref{tab:outro_avllm}.} with AVCD, 
contrastive decoding improves performance, and 
integrating OutRo yields further gains. Since VCD is 
not applicable to Omni-LLMs, we additionally evaluate VCD 
on VideoLLaMA3$_{7\mathrm{B}}$~\cite{zhang2025videollama} and observe the same 
trend, showing that OutRo remains effective alongside 
advanced decoding strategies.

\begin{figure}[t]
\centering
\begin{minipage}[t]{0.48\linewidth}
    \centering
    \includegraphics[width=0.95\linewidth]{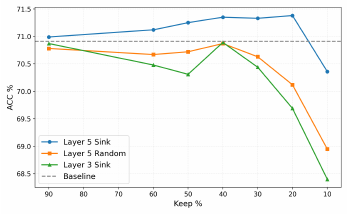}
    \vspace{-3mm}
    \captionof{figure}{\textbf{Token pruning using sink queries.} Pruning tokens by sink-query attention scores improves performance, suggesting sink queries attend to informative modality tokens.}
    \label{fig:supp_sink_prune}
\end{minipage}
\hfill
\begin{minipage}[t]{0.48\linewidth}
    \vspace{-108pt}
    \centering
    \scriptsize
    \setlength{\tabcolsep}{10pt}
    \captionof{table}{\textbf{Sink-query token pruning.} 
    Pruning tokens based on sink-query attention reduces latency as fewer tokens are retained, while performance improves and reaches its peak when 20\% of the tokens are kept.
    }
    \label{tab:latency_prune}
    \vspace{-2mm}
    \resizebox{\linewidth}{!}{
    \begin{tabular}{l c c c}
    \toprule
    \textbf{Keep \%} &
    \makecell{\textbf{Latency$\downarrow$}\\\textbf{(sec/token)}} &
    \textbf{Rel.$\downarrow$} &
    \textbf{Acc.$\uparrow$} \\
    \midrule
    \rowcolor{gray!8} Baseline & 0.996 & 1.00 & 70.91 \\
    60 & 0.904 & 0.91 & 71.12 \\
    50 & 0.864 & 0.87 & 71.25 \\
    40 & 0.824 & 0.83 & 71.35 \\
    30 & 0.784 & 0.79 & 71.33 \\
    20 & 0.749 & 0.75 & 71.38 \\
    10 & 0.714 & 0.72 & 70.36 \\
    \bottomrule
    \end{tabular}}
\end{minipage}
\vspace{-20pt}
\end{figure}

\subpara{Sink queries encode informative directions.}
We analyse whether sink queries attend to informative 
modality tokens by performing token pruning based on 
attention scores between the sink query and modality 
tokens on AVHBench. As shown in \cref{fig:supp_sink_prune}, 
pruning tokens by sink-query attention at Layer~5, where 
OutRo applies sink enhancement, improves performance, 
suggesting that the sink query at this layer already 
encodes a structured direction identifying informative 
tokens. Detailed quantitative results are 
in~\cref{tab:latency_prune}. In contrast, random pruning 
degrades performance, and pruning at Layer~3, where sink 
tokens first emerge, performs even worse, indicating that 
this structured direction only emerges at appropriate 
layers where sufficient context has been aggregated.

\subpara{Omitted details.}
Additional ablation studies and analyses are provided 
in~\cref{supp:analysis_outro} of our appendix.

%% file: tables/outro_avllm.tex
\begin{table*}[t]
\caption{\textbf{Performance comparison of \textit{OutRo}.}
OutRo consistently improves performance across models and benchmarks, demonstrating its generality.}
\vspace{-2mm}
\centering
\setlength{\tabcolsep}{10pt}
\resizebox{\textwidth}{!}{%
\begin{tabular}{lccccccc}
\toprule
\multirow{2}{*}{\textbf{Models}} & 
\multicolumn{4}{c}{\textbf{Audio-Visual QA}} & 
\multicolumn{3}{c}{\textbf{Visual-Only QA}} \\ 
\cmidrule(lr){2-5} \cmidrule(lr){6-8}
& OmniBench & AVUT & AVHBench & DailyOmni 
& Holmes & MME & ActivityNet  \\ 
\midrule
Qwen2.5-Omni$_{3\mathrm{B}}$ & 41.33 & 61.79 & 70.91 & 55.38 & 43.54 & 58.00 & 41.82 \\
\rowcolor{aliceblue}
\textbf{\textit{\texttt{+}OutRo}} & \textbf{42.03} & \textbf{62.87} & \textbf{71.67} & \textbf{55.56} & 43.54 & \textbf{58.11} & \textbf{41.86} \\
\midrule
Qwen2.5-Omni$_{7\mathrm{B}}$ & 47.37 & 65.79 & 71.60 & 54.14 & 47.46 & 62.44 & 43.59 \\
\rowcolor{aliceblue}
\textbf{\textit{\texttt{+}OutRo}} & \textbf{48.25} & \textbf{66.57} & \textbf{73.78} & \textbf{54.64} & \textbf{47.63} & \textbf{62.96} & \textbf{44.01} \\
\midrule
video-SALMONN2\texttt{+}$_{3\mathrm{B}}$ & 36.69 & 56.73 & 62.07 & 51.71 &  42.68 &  73.67 & 44.31 \\
\rowcolor{aliceblue}
\textbf{\textit{\texttt{+}OutRo}} & \textbf{36.95} & \textbf{57.31} & \textbf{62.33} & \textbf{52.05} & \textbf{42.90} & \textbf{73.89} & \textbf{44.41} \\
\bottomrule
\end{tabular}
}
\vspace{-4mm}
\label{tab:outro_avllm}
\end{table*}

%% file: tables/decoding_layench_compat_cd.tex
\begin{table}[h]
\centering
\scriptsize
\begin{minipage}[t]{0.48\linewidth}
\centering
\setlength{\tabcolsep}{8pt}
\renewcommand{\arraystretch}{0.95}
\captionof{table}{\textbf{Decoding efficiency.} OutRo introduces only $1.11\times$ overhead with negligible memory overhead compared to Flash-attn.}
\label{tab:latency}
\resizebox{\linewidth}{!}{
\begin{tabular}{l c c c}
\toprule
\textbf{Decoding} &
\makecell{\textbf{GPU Mem.$\downarrow$} \\ \textbf{(GiB)}} &
\makecell{\textbf{Latency$\downarrow$} \\ \textbf{(sec/token)}} &
\textbf{Rel.$\downarrow$} \\
\midrule
\rowcolor{gray!8} Flash-attn & 11.48 & 0.98 & 1.00 \\
VCD~\cite{leng2024mitigating} & 13.54 & 1.98 & 2.02 \\
\rowcolor{aliceblue}
\textbf{\textit{OutRo}} & \textbf{11.48} & \textbf{1.09} & \textbf{1.11} \\
\midrule
\rowcolor{gray!8}
Eager & 17.87 & 2.00 & 2.04 \\
VAR~\cite{kang2025see} & 19.02 & 2.71 & 2.77 \\
AVCD~\cite{jung2025avcd} & 27.65 & 7.88 & 8.04 \\
\bottomrule
\end{tabular}}
\end{minipage}
\hfill
\begin{minipage}[t]{0.48\linewidth}
\centering
\setlength{\tabcolsep}{8pt}
\renewcommand{\arraystretch}{1.0}
\captionof{table}{\textbf{Contrastive decoding.} OutRo is compatible with CD, yielding additional gains.}
\label{tab:ablation_cd}
\resizebox{\linewidth}{!}{
\begin{tabular}{l c c}
\toprule
\textbf{Qwen2.5-Omni~\cite{xu2025qwen25omnitechnicalreport}} & OmniBench & DailyOmni \\
\midrule
\rowcolor{gray!8} Baseline & 38.79 & 54.55 \\
AVCD~\cite{jung2025avcd} & 39.58 & 55.56 \\
\rowcolor{aliceblue}
\textbf{\textit{\texttt{+}OutRo}} & \textbf{40.81} & \textbf{55.89} \\
\toprule
\textbf{VideoLLaMA3~\cite{zhang2025videollama}} & Holmes & MME \\
\midrule
\rowcolor{gray!8} Baseline & 41.15 & 61.78 \\
VCD~\cite{leng2024mitigating} & 42.90 & 62.00 \\
\rowcolor{aliceblue}
\textbf{\textit{\texttt{+}OutRo}} & \textbf{43.22} & \textbf{62.22} \\
\bottomrule
\end{tabular}}
\end{minipage}
\vspace{-10pt}
\end{table}

%% file: sections/supp.tex
\clearpage

\appendix

\renewcommand{\theHsection}{appendix.\Alph{section}}
\renewcommand{\theHsubsection}{appendix.\Alph{section}.\arabic{subsection}}

\counterwithin{figure}{section}
\counterwithin{table}{section}

\renewcommand{\thefigure}{\thesection.\arabic{figure}}
\renewcommand{\thetable}{\thesection.\arabic{table}}

\makeatletter
\providecommand{\authcount}[1]{}
\makeatother

\begin{center}
    \Large \textbf{On the Nature of Attention Sink \\ that Shapes Decoding Strategy in Omni-LLMs} \\

    \vspace{0.5em}
    \large \textbf{-- Appendix --} \\ 
    \vspace{1.5em} 
\end{center}

\startcontents[supp]
\section*{Contents}
\hypersetup{linkcolor=black}
\printcontents[supp]{l}{1}{
\setcounter{tocdepth}{2}
\setlength{\baselineskip}{1.4\baselineskip}
}

\clearpage

\section{Analysis for Sink Identification}
This section provides additional analyses of the sink identification criterion discussed in Sec.~3 (Q1), including statistics of sink and outlier tokens, empirical results, and qualitative examples. All analyses are conducted on Qwen2.5-Omni.

\subsection{Sink and Outlier Statistics}
\label{supp:sink_stats}
\input{tables/supp_stat_outlier_sink}
We report the top 30 activation dimensions in~\cref{tab:stat_outlier_sink}. 
A clear overlap is observed between the dimensions identified as outliers and those detected as sink dimensions. 
In particular, prominent sink dimensions such as 458 and 2570 exhibit large activations across a wide range of tokens.

Under the $\Phi_{\text{VLM}}$ criterion, these large activations cause many tokens to be classified as sinks, resulting in a substantial fraction of tokens being labelled as sink tokens, including semantically meaningful visual tokens.
This suggests that $\Phi_{\text{VLM}}$ tends to over-identify sinks in multimodal settings.

In contrast, the $\Phi_{\text{LLM}}$ criterion identifies only a single token as the sink and consistently assigns the sink index to a structurally defined token, leading to a more selective and stable identification of the sink token.

\subsection{Empirical Results}
\label{supp:sink_empirical}
\input{tables/supp_sink_remove}
To further validate the sink identification criteria, we conduct a controlled intervention by deactivating the sink dimensions $D_{\text{sink}}$ (\cref{tab:supp_sink_remove}).

First, when the dimensions $D_{\text{sink}}$ are disabled for all tokens, the model performance collapses across all tasks. 
Similarly, when applying the LLM-based criterion $\Phi_{\text{LLM}}$, disabling $D_{\text{sink}}$ only at the token identified as a sink also leads to a severe performance drop, suggesting that the token detected by $\Phi_{\text{LLM}}$ functions as a stabilising anchor in the model's computation. In contrast, when a token identified by the VLM-based criterion $\Phi_{\text{VLM}}$ is randomly selected and its $D_{\text{sink}}$ dimensions are disabled, the model performance remains close to the baseline.
Furthermore, even when $D_{\text{sink}}$ are disabled for all tokens detected by $\Phi_{\text{VLM}}$ except the token identified by $\Phi_{\text{LLM}}$, the performance remains largely unchanged.

These results suggest that the token identified by $\Phi_{\text{LLM}}$ plays a more critical role in the model's inference process than those detected by $\Phi_{\text{VLM}}$.

\subsection{Additional Qualitative Examples}
\label{supp:q1_qualitative}

\begin{figure}[t]
    \centering
    \includegraphics[width=0.8\linewidth]{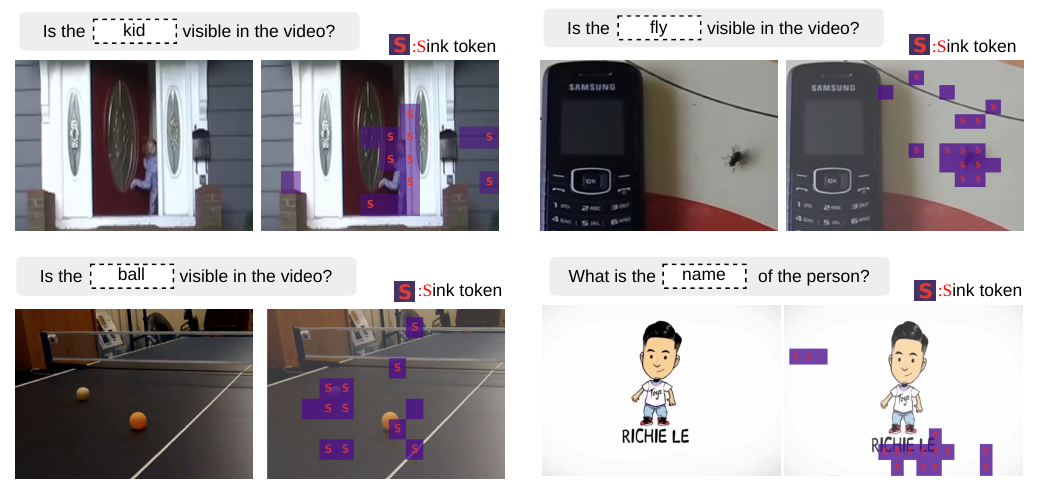}
    \vspace{-3mm}
    \caption{\textbf{Additional qualitative examples.} Each example visualises attention scores for query tokens (\textit{e.g.,} kid, fly, ball, name) on AVHBench videos.}
    \label{fig:q1_qualitative}
    \vspace{-5mm}
\end{figure}

We provide further qualitative examples illustrating the misbehaviour of the visual attention sink criterion $\Phi_{\text{VLM}}$ when applied to Omni-LLMs. In~\cref{fig:q1_qualitative}, we observe the same pattern where a large fraction of tokens are classified as sinks, including tokens corresponding to semantically meaningful visual content.

\subsection{Sink Identification on video-SALMONN2+}
\label{supp:sink_id_salmonn}

\begin{figure}[t]
    \centering
    \includegraphics[width=0.8\linewidth]{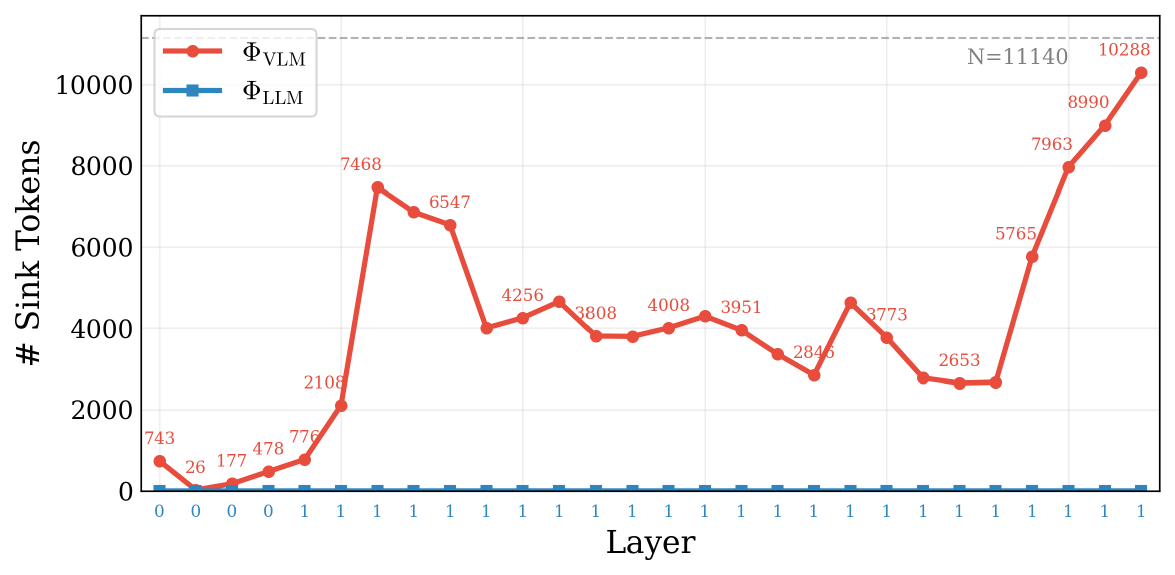}
    \caption{\textbf{Number of sink tokens across layers under $\Phi_{\text{VLM}}$ and $\Phi_{\text{LLM}}$ on video-SALMONN2+.} $\Phi_{\text{VLM}}$ rapidly over-identifies sink tokens as depth increases, eventually classifying nearly all tokens ($N=11140$) as sinks in the final layers. In contrast, $\Phi_{\text{LLM}}$ consistently identifies only a sparse and stable set of sink tokens throughout all layers.}
    \label{fig:supp_salmonn_sink}
\end{figure}

We provide additional sink identification results on video-SALMONN2+ to support the findings in Q1. As shown in \cref{fig:supp_salmonn_sink}, the same pattern observed in Qwen2.5-Omni holds: $\Phi_{\text{VLM}}$ collapses and over-identifies the majority of tokens as sinks in deeper layers, while $\Phi_{\text{LLM}}$ maintains a sparse and stable identification throughout. This confirms that $\Phi_{\text{LLM}}$ generalises reliably across different Omni-LLM architectures.

\input{tables/vlm_tau_sensitive}
\subsection{Sensitivity to the Sink Threshold}
We examine whether the over-identification of sink tokens under $\Phi_{\mathrm{VLM}}$ can be resolved by adjusting its threshold $\tau$. We vary $\tau \in \{20,25,30\}$ and measure the proportion of tokens identified as sinks across four audio-visual benchmarks on Qwen2.5-Omni 3B and 7B. Increasing $\tau$ substantially reduces the number of detected sinks, but its effect varies markedly across model scales. On the 3B model, the proportion falls from 10.02--15.33\% at $\tau=20$ to 0.03--0.13\% at $\tau=30$. In contrast, the same threshold of $\tau=30$ still identifies 5.98--8.27\% of tokens as sinks on the 7B model, compared with 24.53--33.32\% at $\tau=20$. These results indicate that $\Phi_{\mathrm{VLM}}$ is highly sensitive to threshold selection, and that a threshold yielding sparse detections at one model scale does not necessarily transfer to another. In comparison, $\Phi_{\mathrm{LLM}}$ consistently identifies a single structurally defined sink token without model-specific threshold tuning.

\section{Analysis of OutRo}
\label{supp:analysis_outro}

In this section, we further investigate the design choices and the analysis of OutRo. Unless otherwise specified, all analyses are conducted on Qwen2.5-Omni.

\subpara{Gated rotation.}
We also investigate using the sink score directly as a 
rotation criterion. Although tuned thresholds 
(\textit{e.g.,} 0.7) can yield larger improvements over 
the baseline (\cref{tab:ablation_sink_score}), this 
approach requires access to attention maps or pre-computed 
sink statistics, introducing overhead and hyper-parameter 
tuning. In contrast, the proxy based on cosine similarity 
(\cref{eq:proxy}) achieves performance gains while 
preserving FlashAttention
compatibility and computational efficiency.

We therefore adopt gated rotation (\cref{eq:gate}). 
Instead of uniform rotation, we modulate head outputs 
with a Tanh gate, enabling adaptive control over the 
rotation strength. This outperforms plain rotation, 
indicating that controlled modulation is empirically 
more effective (\cref{tab:ablation_gate}).

\subpara{Sink enhancement.}
The timing of sink enhancement is critical. If applied 
too early, sink tokens have not yet aggregated sufficient 
global information and their queries remain weakly 
structured, leading to limited or degraded performance 
(\cref{tab:ablation_enh}). This aligns with prior 
findings that transformer layers exhibit an early stage 
of modality structuring before deeper integration, 
typically spanning the first seventh of 
layers~\cite{jung2025fork, yu2025multimodal, wei2024phase, 
khaki2025sparsevila}. We also compare applying enhancement 
once at this depth with applying it at every subsequent 
layer. A single application performs better, indicating 
that intervention at the transition depth is sufficient, 
while repeated enhancement introduces redundancy.

\begin{table}[t]
\centering
\scriptsize
\begin{minipage}[t]{0.32\linewidth}
\centering
\captionof{table}{\textbf{Sink score vs. cosine similarity.} Cosine gating improves over the baseline.}
\label{tab:ablation_sink_score}
\setlength{\tabcolsep}{20pt}
\begin{tabular}{l c}
\toprule
\textbf{Criterion} & \textbf{Acc.} \\
\midrule
\rowcolor{gray!8} Baseline & 70.91 \\
0.6 & 72.46 \\
0.7 & \textbf{72.82} \\
0.8 & 72.72 \\
0.9 & 71.46 \\
Cosine & 71.33 \\
\bottomrule
\end{tabular}
\end{minipage}
\hfill
\begin{minipage}[t]{0.32\linewidth}
\centering
\captionof{table}{\textbf{Gate study.} Tanh gate outperforms uniform and polynomial alternatives on AVHBench.}
\label{tab:ablation_gate}
\setlength{\tabcolsep}{20pt}
\begin{tabular}{l c}
\toprule
\textbf{Gate} & \textbf{Acc.} \\
\midrule
\rowcolor{gray!8} Baseline & 70.91 \\
Uniform & 71.33 \\
Quadratic & 70.78 \\
Cubic & 70.67 \\
Tanh & \textbf{71.44} \\
\bottomrule
\end{tabular}
\end{minipage}
\hfill
\begin{minipage}[t]{0.32\linewidth}
\centering
\captionof{table}{\textbf{Enhancement layer ablation.} A single application at L5 outperforms repeated enhancement on AVHBench.}
\label{tab:ablation_enh}
\setlength{\tabcolsep}{20pt}
\begin{tabular}{l c}
\toprule
\textbf{Layer} & \textbf{Acc.} \\
\midrule
\rowcolor{gray!8} Baseline & 70.91 \\
L3 & 70.67 \\
L4 & 70.94 \\
L5 & 71.08 \\
L5 (Once) & \textbf{71.25} \\
\bottomrule
\end{tabular}
\end{minipage}
\end{table}



\subsection{Directional Proxy for Sink Score}
\label{supp:proxy_sink}

\begin{figure}[t]
    \centering
    \includegraphics[width=0.8\linewidth]{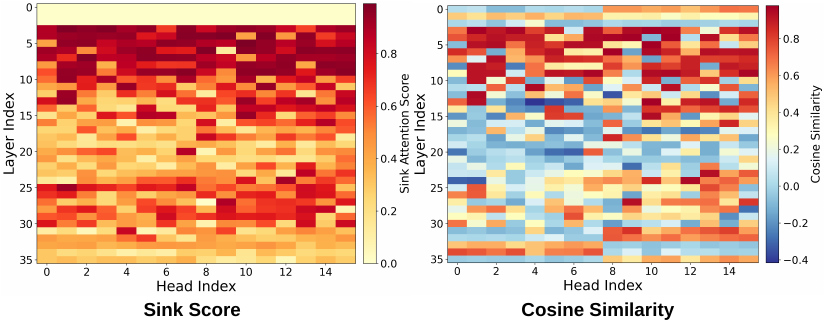}
    \vspace{-3mm}
    \caption{\textbf{Sink score vs. cosine similarity.} Attention to sink tokens is strongly correlated with the cosine similarity between head outputs and the sink value direction.}
    \label{fig:fig9}
    \vspace{-4mm}
\end{figure}

\cref{fig:fig9} shows a strong positive correlation between attention to sink tokens and the cosine similarity between each head output vector and the sink value direction, with an average Pearson correlation of approximately $0.7$. This indicates that directional alignment serves as an effective proxy for sink score, enabling a more efficient formulation used in our method.

\subsection{Attention Redistribution After Rotation}
\label{supp:rotate_attn}

We analyse how rotation affects the distribution of 
attention across token types. For each model, we measure 
the change in total attention mass allocated to sink, 
video, and audio tokens before and after rotation. To 
account for differences in the number of tokens per 
modality, baseline attention values are normalised by 
the number of modality tokens.

As shown in \cref{tab:rotate_attention_delta_gamma1}, 
rotation consistently reduces attention to sink tokens 
across all models, with the freed attention redistributing 
toward modality tokens. Notably, the redistribution is 
not uniform across modalities: in Qwen2.5-Omni~\cite{xu2025qwen25omnitechnicalreport} and 
video-SALMONN2+~\cite{tang2025video}, where audio tokens originally receive 
more attention per token than video tokens, the increase 
is larger for video; in VideoLLaMA2~\cite{cheng2024videollama}, where video tokens 
receive more attention per token, the increase is larger 
for audio. This suggests that rotation normalises the 
attention imbalance across modalities, directing the 
released capacity toward the modality that the model 
originally attended to less.

\begin{table}[t]
\centering
\small
\caption{\textbf{Attention redistribution after rotation.} 
Attention to video/audio values are baseline attention mass 
normalised by the number of modality tokens. Deltas show 
the change in total attention mass after rotation. 
Rotation reduces sink attention and preferentially 
increases attention toward the modality that was 
originally attended to less.}
\begin{tabular}{lccrrr}
\toprule
Model & Attn. to video / tok & Attn. to audio / tok & $\Delta$ Sink & $\Delta$ Video & $\Delta$ Audio \\
\midrule
Qwen2.5-Omni~\cite{xu2025qwen25omnitechnicalreport} & 2.20e-04 & \textbf{2.87e-04} & -0.0176 & \textbf{+0.0127} & +0.0022 \\
video-SALMONN2+~\cite{tang2025video} & 6.45e-05 & \textbf{3.64e-04} & -0.0095 & \textbf{+0.0088} & +0.0005 \\
VideoLLaMA2~\cite{cheng2024videollama} & \textbf{3.22e-04} & 2.15e-04 & -0.0234 & +0.0103 & \textbf{+0.0123} \\
\bottomrule
\end{tabular}
\label{tab:rotate_attention_delta_gamma1}
\end{table}

\subsection{Sink Enhancements Sources}
\input{tables/supp_enhancement_sources}

We analyse different sources for applying attention between the sink query and modality keys on AVHBench. In addition to the internal features used in OutRo, we also validate encoder inputs by directly applying cross-attention to the visual and audio modalities. As shown in~\cref{tab:enhancement_sources}, using encoder inputs is less effective than using internal features. Removing the text modality and applying cross-attention only to the visual and audio inputs also does not improve performance.

\subsection{Results on VLMs}
\label{supp:vlm_results}

\cref{tab:outro_vlm} reports results on visual-only LLMs.
Across Qwen2.5-VL, Qwen3-VL, and VideoLLaMA3, OutRo 
improves performance on visual-only QA benchmarks.
For example, Qwen3-VL achieves a gain of \texttt{+}0.89 
on VideoMME, and VideoLLaMA3 improves by \texttt{+}1.47 
on VideoHolmes. These results demonstrate that OutRo 
generalises robustly across diverse VLMs regardless of 
modality composition, improving visual reasoning.

\begin{table}[t]
\centering
\setlength{\tabcolsep}{13pt}
\renewcommand{\arraystretch}{1.0}
\caption{Performance of \textit{\textbf{OutRo}} on VLMs.}
\label{tab:outro_vlm}
\resizebox{0.7\linewidth}{!}{
\begin{tabular}{lccc}
\toprule
\textbf{Models} & Holmes & MME & ActivityNet \\
\midrule
Qwen2.5-VL$_{3\mathrm{B}}$~\cite{bai2025qwen25vltechnicalreport} & 41.26 & 57.89 & \textbf{47.02} \\
\rowcolor{aliceblue}
\textbf{\textit{\texttt{+}OutRo}} & \textbf{42.41} & \textbf{58.33} & 46.87 \\
\midrule
Qwen3-VL$_{8\mathrm{B}}$~\cite{bai2025qwen3} & 46.65 & 67.44 & 46.65 \\
\rowcolor{aliceblue}
\textbf{\textit{\texttt{+}OutRo}} & \textbf{46.76} & \textbf{68.33} & \textbf{46.85} \\
\midrule
VideoLLaMA3$_{7\mathrm{B}}$~\cite{zhang2025videollama} & 41.15 & 61.78 & 49.32 \\
\rowcolor{aliceblue}
\textbf{\textit{\texttt{+}OutRo}} & \textbf{42.62} & \textbf{62.56} & \textbf{49.92} \\
\bottomrule
\end{tabular}}
\end{table}

\subsection{Effect of Rotation Strength $\gamma$}

\begin{figure}[t]
\centering

\begin{subfigure}{0.48\linewidth}
\centering
\includegraphics[width=\linewidth]{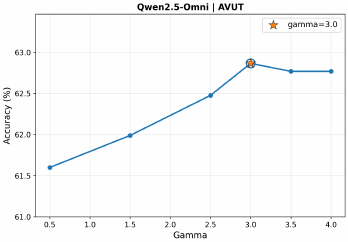}
\vspace{-4mm}
\label{fig:gamma1}
\end{subfigure}
\hfill
\begin{subfigure}{0.48\linewidth}
\centering
\includegraphics[width=\linewidth]{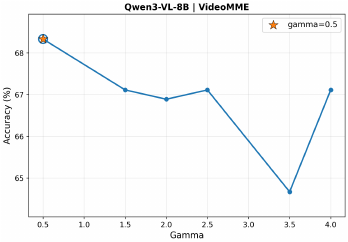}
\vspace{-4mm}
\label{fig:gamma2}
\end{subfigure}

\vspace{-2mm}

\begin{subfigure}{0.48\linewidth}
\centering
\includegraphics[width=\linewidth]{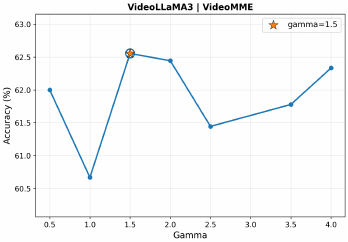}
\vspace{-4mm}
\label{fig:gamma3}
\end{subfigure}
\hfill
\begin{subfigure}{0.48\linewidth}
\centering
\includegraphics[width=\linewidth]{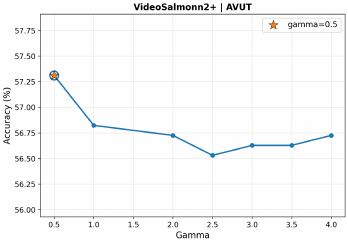}
\vspace{-4mm}
\label{fig:gamma4}
\end{subfigure}

\vspace{-3mm}
\caption{\textbf{Effect of different $\gamma$ values.} Optimal $\gamma$ values vary across models: $3.0$ for Qwen2.5, 
$0.5$ for Qwen3-VL and video-SALMONN2\texttt{+}, and $1.5$ for VideoLLaMA3.}
\label{fig:gamma_ablation}
\vspace{-4mm}

\end{figure}
We analyse the sensitivity of OutRo to the rotation strength parameter $\gamma$, which controls the magnitude of modulation applied to attention head outputs. \cref{fig:gamma_ablation} reports results across representative Omni-LLMs and VLMs on the benchmarks. We vary $\gamma$ in the range $[0.5, 4.0]$ and observe stable performance trends within each model, with moderate variation across architectures. Based on these results, we select the optimal $\gamma$ for each model in our experiments.

\input{tables/rotation_vlm_top1}
\subsection{Rotation with Different Sink Identification Criteria}
\label{sec:sink_criterion_rotation}
We examine whether a stricter selection under $\Phi_{\mathrm{VLM}}$ recovers sink tokens that provide an effective direction for rotation. We retain either the top 1\% of tokens or the single highest-scoring token under $\Phi_{\mathrm{VLM}}$, and use their value representations to construct the rotation direction. We isolate the rotation component of OutRo and keep the remaining intervention unchanged, allowing us to assess the effect of the sink identification criterion. Experiments are conducted on Qwen2.5-Omni 3B and 7B across four audio-visual benchmarks, using stochastic decoding with temperature 0.7 and top-$p$ 0.9. We report the mean and standard deviation over five seeds.

As shown in Tab.~\ref{tab:sink_criterion_rotation}, both selections under $\Phi_{\mathrm{VLM}}$ degrade performance relative to the baseline across all eight model--benchmark settings. Selecting only the highest-scoring token generally reduces the degradation compared with the top 1\% selection, but does not recover baseline performance. In contrast, rotation using $\Phi_{\mathrm{LLM}}$ outperforms both alternatives in every setting and improves over the baseline in six of the eight settings. These results suggest that the distinction between the criteria is not explained solely by the number of selected tokens: restricting $\Phi_{\mathrm{VLM}}$ to a small subset does not recover a comparably effective rotation direction.

\input{tables/image_text_llm}
\subsection{Generalisation to Image--Text and Text-Only Settings}
\label{sec:image_text_generalisation}

We examine whether the benefits of OutRo extend beyond audio-visual inputs. We first evaluate Qwen2.5-Omni 3B under image--text inputs on POPE~\cite{li2023evaluating} and text-only inputs on GSM8K~\cite{cobbe2021training}. Both experiments use the full evaluation sets with greedy decoding. As shown in Tab.~\ref{tab:image_text_generalisation}, OutRo improves performance in both settings, indicating that its effectiveness does not depend on the simultaneous presence of video and audio.

We further evaluate OutRo on two text-only LLMs, Qwen2.5-7B-Instruct~\cite{qwen2025qwen25technicalreport} and Llama-3.1-8B-Instruct~\cite{grattafiori2024llama}, using GSM8K. We report mean accuracy and standard deviation over five seeds, with correctness determined by matching the final numeric answer. The rotation strength $\gamma$ is selected on 100 validation samples, yielding $\gamma=3$ for Qwen and $\gamma=1$ for Llama. OutRo improves performance on both models, suggesting that the intervention also applies beyond Omni-LLM backbones. Together, these results support the applicability of OutRo across different input configurations and model families.

\input{tables/random_rotation}
\subsection{Rotation toward Alternative Directions}
\label{sec:random_direction_rotation}

We examine whether the benefit of rotation depends on the sink value direction or can be obtained using alternative directions. We replace the sink value direction with the value representation of a randomly selected video, audio, or text token, and additionally evaluate a head-mean direction. All variants retain the same gated rotation and norm-preserving normalisation. We use the rotation component alone, without mask relaxation, to isolate the effect of the chosen direction. Experiments are conducted on Qwen2.5-Omni 3B across four audio-visual benchmarks, using stochastic decoding with temperature 0.7 and top-$p$ 0.9. We report the mean and standard deviation over five seeds.

As shown in Tab.~\ref{tab:random_direction_rotation}, all alternative directions degrade performance relative to the baseline across the four benchmarks. Random token directions reduce accuracy regardless of the selected modality, while the head-mean direction produces the largest degradation in every setting. In contrast, rotation toward the sink value direction consistently improves performance. Since all variants preserve the original head-output norm, these differences cannot be explained by changes in output magnitude alone. These results support the importance of the sink value direction, rather than rotation toward an arbitrary token or the head-mean direction.

\section{Discussion}

\subsection{Broader Impacts}
This work provides a better understanding of how 
attention sinks operate in Omni-LLMs, offering insights 
into the internal mechanisms of multimodal transformers. 
Building on this analysis, OutRo improves audio-visual 
reasoning in a training-free manner, which could benefit 
real-world applications such as assistive technologies, 
video summarisation, and multimodal dialogue systems.

As a training-free method that does not modify model 
weights or training data, OutRo does not introduce new 
risks beyond those already present in the underlying 
models. The potential for misuse is inherited from the 
base Omni-LLMs rather than introduced by our approach.

\subsection{Limitations}
The rotation strength parameter $\gamma$ requires manual specification and its optimal value may vary across model architectures. In addition, although OutRo improves performance across tasks, the magnitude of improvement is modest. Developing strategies to reduce the calibration effort for $\gamma$ and further enhance its effectiveness remains an important direction for future work.

\section{Algorithm of OutRo}
\input{algorithm/algorithm}
\cref{alg:outro} summarises the OutRo procedure applied at each decoder layer. 
Given the hidden states, we first identify sink tokens using the sink criterion $\Phi_{\mathrm{LLM}}$. 
We then compute the attention outputs and estimate the sink value direction by averaging the value vectors of the detected sink tokens.

For non-sink positions, head outputs are modulated through gated rotation according to their directional alignment with the sink value direction, controlled by the rotation strength $\gamma$. 
The updated outputs are rescaled to preserve their original magnitude.
For sink positions, we apply mask relaxation at a selected layer, allowing sink queries to attend beyond the causal mask and thereby aggregate contextual information. 
The resulting outputs are then combined across heads and projected to produce the hidden states that are fed into the subsequent FFN layer.

\section{Qualitative Results}

\begin{figure}[h]
\centering

\begin{subfigure}{\linewidth}
\centering
\includegraphics[width=\linewidth]{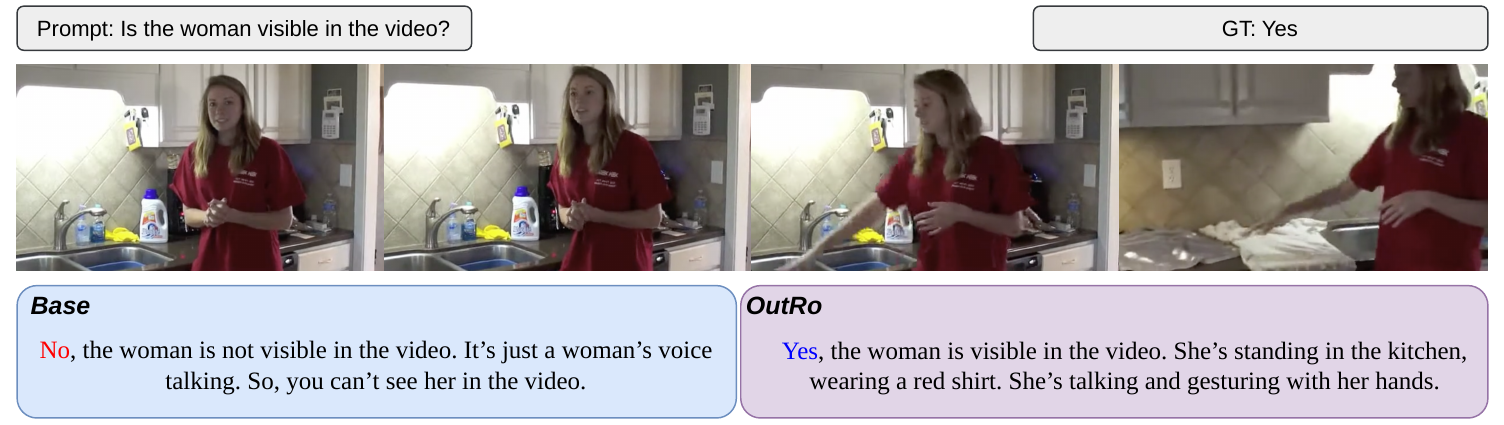}
\vspace{-4mm}
\caption{}
\end{subfigure}

\vspace{2mm}

\begin{subfigure}{\linewidth}
\centering
\includegraphics[width=\linewidth]{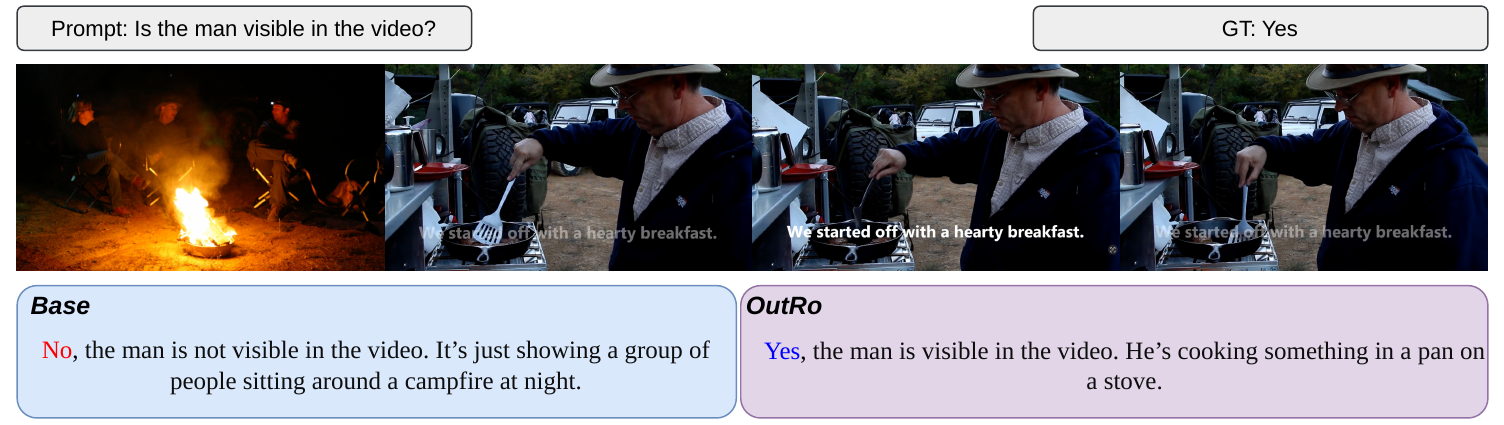}
\vspace{-4mm}
\caption{}
\end{subfigure}

\vspace{2mm}

\begin{subfigure}{\linewidth}
\centering
\includegraphics[width=\linewidth]{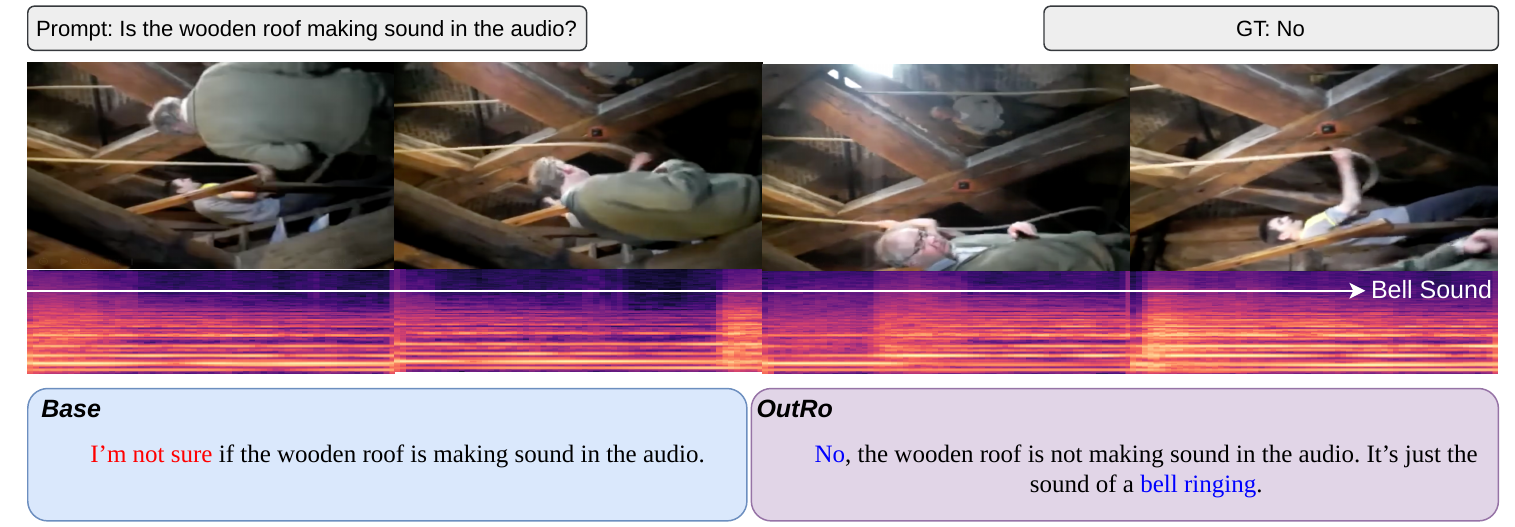}
\vspace{-4mm}
\caption{}
\end{subfigure}

\vspace{2mm}

\begin{subfigure}{\linewidth}
\centering
\includegraphics[width=\linewidth]{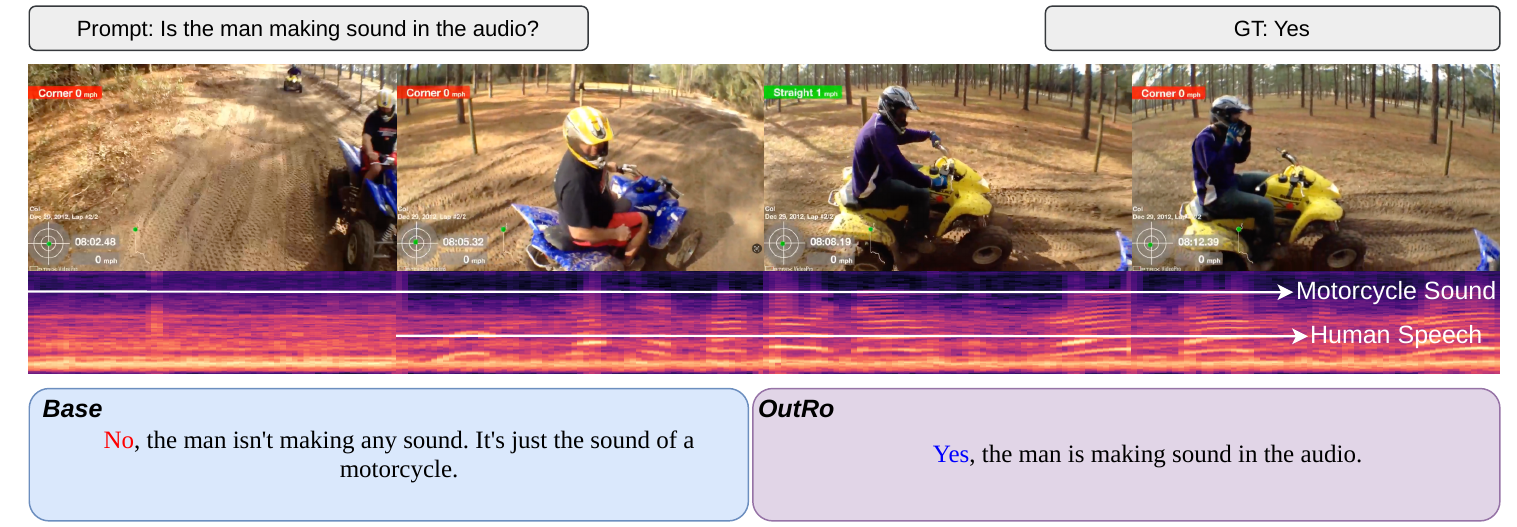}
\vspace{-4mm}
\caption{}
\end{subfigure}

\vspace{-3mm}
\caption{\textbf{Qualitative results on video-QA on AVHBench.} Compared with the baseline, OutRo produces answers that better align with visual and audio-visual evidence in the video, leading to more accurate multimodal reasoning.}
\vspace{-4mm}

\label{fig:qualitative1}
\end{figure}

\begin{figure}[h]
\centering

\begin{subfigure}{\linewidth}
\centering
\includegraphics[width=\linewidth]{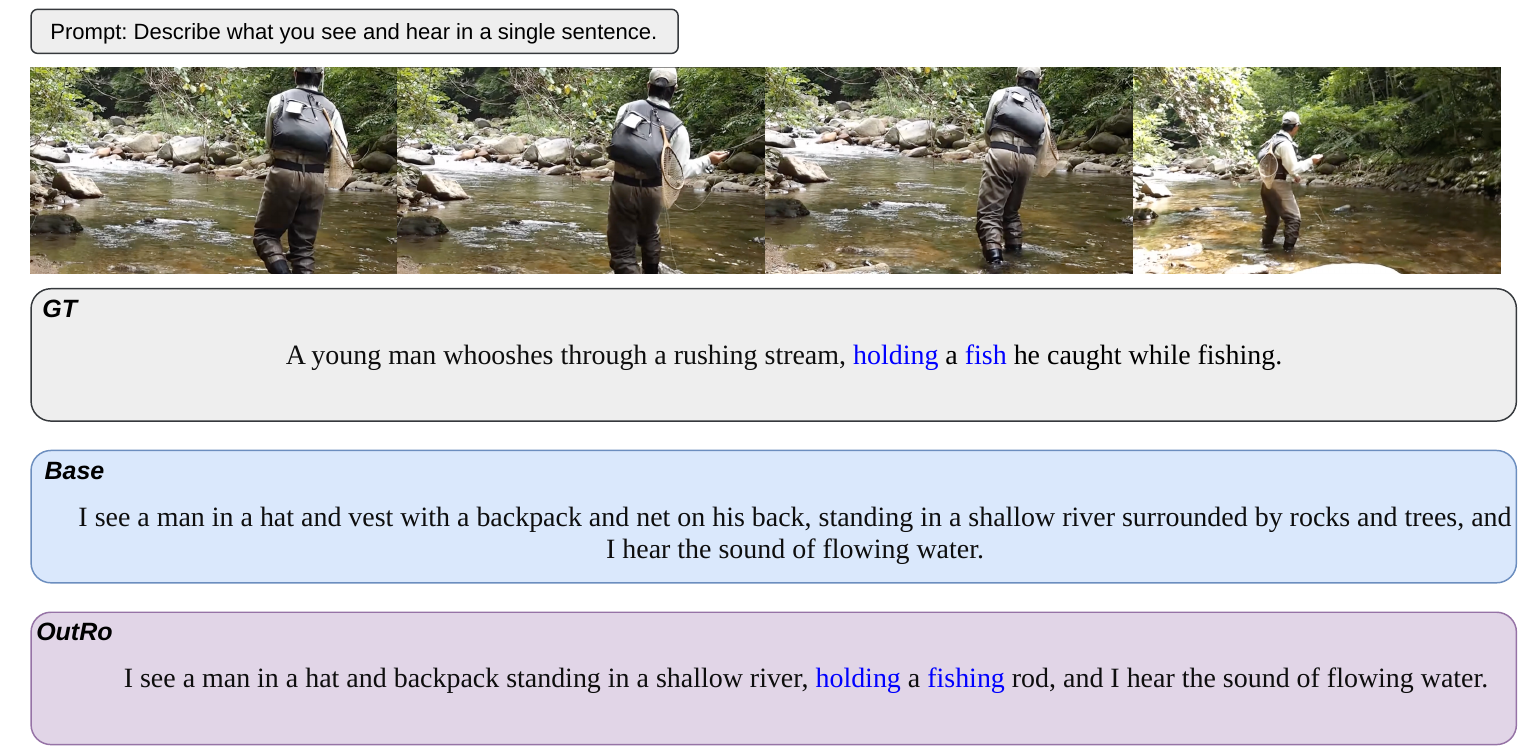}
\vspace{-4mm}
\caption{}
\end{subfigure}

\vspace{2mm}

\begin{subfigure}{\linewidth}
\centering
\includegraphics[width=\linewidth]{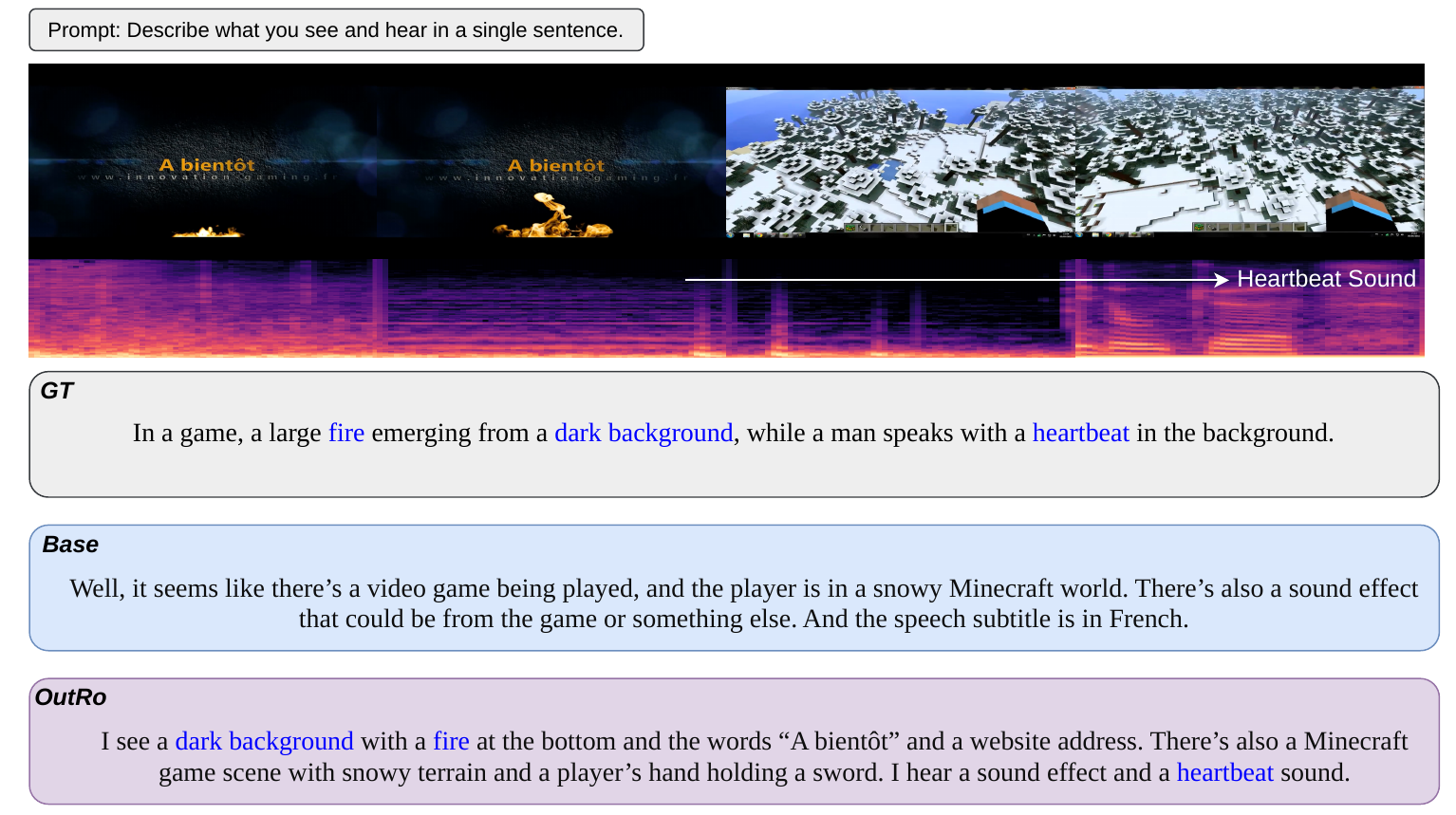}
\vspace{-4mm}
\caption{}
\end{subfigure}

\vspace{-3mm}
\caption{\textbf{Qualitative results for video captioning on AVHBench.} OutRo produces captions grounded in both visual and audio-visual cues, whereas the baseline often fails to capture such multimodal details.}
\vspace{-4mm}

\label{fig:qualitative2}
\end{figure}

We present qualitative examples illustrating the behaviour of OutRo across different multimodal models. Examples on video-QA tasks are shown in~\cref{fig:qualitative1}, while video captioning results are presented in~\cref{fig:qualitative2}. These qualitative observations are consistent with the quantitative improvements, where captioning performance on AVHBench increases from $3.01$ to $3.11$.

%% file: tables/supp_stat_outlier_sink.tex
\begin{table*}[t]
\caption{\textbf{Dimensions identified as outliers or sinks.} Under the $\Phi_{\text{VLM}}$ criterion, sink dimensions (\textit{i.e.,} 458 and 2570) cause many tokens to be classified as sinks.}
\centering
\scriptsize
\begin{tabular}{c c  c c c  c c c  c c c}
\toprule
 &  & \multicolumn{3}{c}{\textbf{Outlier}} & \multicolumn{3}{c}{\textbf{Sink $\Phi_{\text{VLM}}$}} & \multicolumn{3}{c}{\textbf{Sink $\Phi_{\text{LLM}}$}} \\
\cmidrule(lr){3-5} \cmidrule(lr){6-8} \cmidrule(lr){9-11}
Dim & Activation & Layer(\%) & Token(\%) & Outlier & $\Phi_{\text{VLM}}$ & Token(\%) & Sink & $\Phi_{\text{LLM}}$ & Index & Sink \\
\midrule
458  & 8768.00 & 100.0 & 86.89 & \cmark & 46.00 & 59.83 & \cmark & 19072 & 2 & \cmark \\
2570 & 6272.00 & 100.0 & 89.80 & \cmark & 45.75 & 44.00 & \cmark & 13376 & 2 & \cmark \\
2718 & 2656.00 & 92.9 & 2.68  &        & 46.00 & 59.83 & \cmark & 6208  & 2 & \cmark \\
2730 & 2128.00 & 100.0 & 2.75 &        & 46.00 & 59.83 & \cmark & 4960  & 2 & \cmark \\
3206 & 716.00  & 100.0 & 57.34 & \cmark & 46.00 & 72.68 & \cmark & 1640 & 2 & \cmark \\
3281 & 612.00  & 100.0 & 16.74 & \cmark & 46.00 & 59.83 & \cmark & 1448 & 2 & \cmark \\
1427 & 604.00  & 100.0 & 9.95 & \cmark & 46.00 & 59.83 & \cmark & 1400 & 2 & \cmark \\
1803 & 556.00  & 100.0 & 3.15 &        & 46.00 & 72.68 & \cmark & 1264 & 2 & \cmark \\
1692 & 556.00  & 100.0 & 11.10 & \cmark & 46.00 & 59.83 & \cmark & 1320 & 2 & \cmark \\
3110 & 544.00  & 100.0 & 25.90 & \cmark & 46.00 & 59.83 & \cmark & 1256 & 2 & \cmark \\
2107 & 544.00  & 100.0 & 42.05 & \cmark & 45.75 & 44.00 & \cmark & 1192 & 2 & \cmark \\
1923 & 512.00  & 100.0 & 42.13 & \cmark & 46.00 & 59.83 & \cmark & 1240 & 2 & \cmark \\
32   & 486.00  & 100.0 & 64.11 & \cmark & 46.00 & 59.83 & \cmark & 1168 & 2 & \cmark \\
1451 & 400.00  & 96.4 & 25.95 & \cmark & 46.00 & 59.83 & \cmark & 936  & 2 &  \\
1111 & 362.00  & 89.3 & 8.00  & \cmark & 45.75 & 44.00 & \cmark & 800  & 2 &  \\
3197 & 350.00  & 100.0 & 92.90 & \cmark & 45.75 & 44.00 & \cmark & 764 & 2 &  \\
392  & 247.00  & 96.4 & 25.92 & \cmark & 45.75 & 9.77 & \cmark & 592  & 2 &  \\
3461 & 229.00  & 100.0 & 10.43 & \cmark & 45.75 & 9.77 & \cmark & 548 & 2 &  \\
662  & 212.00  & 100.0 & 24.51 & \cmark & 46.00 & 59.83 & \cmark & 484 & 2 &  \\
1790 & 190.00  & 100.0 & 14.45 & \cmark & 46.00 & 59.83 & \cmark & 456 & 2 &  \\
637  & 184.00  & 89.3 & 6.60  & \cmark & 45.75 & 9.77 & \cmark & 440 & 2 &  \\
1627 & 173.00  & 100.0 & 19.78 & \cmark & 45.75 & 44.00 & \cmark & 382 & 2 &  \\
608  & 169.00  & 100.0 & 32.64 & \cmark & 46.00 & 72.68 & \cmark & 396 & 2 &  \\
162  & 169.00  & 89.3 & 6.09  & \cmark & 1.54  & 0.12 &  & 80  & 120 &  \\
1431 & 160.00  & 89.3 & 28.71 & \cmark & 45.75 & 44.00 & \cmark & 350 & 2 &  \\
143  & 150.00  & 100.0 & 7.67  & \cmark & 45.75 & 9.77 & \cmark & 358 & 2 &  \\
2069 & 148.00  & 89.3 & 6.39  & \cmark & 45.75 & 9.77 & \cmark & 354 & 2 &  \\
2591 & 142.00  & 100.0 & 8.54  & \cmark & 4.50  & 0.12 &  & 68  & 15  &  \\
68   & 137.00  & 89.3 & 6.30  & \cmark & 46.00 & 59.83 & \cmark & 326 & 2 &  \\
882  & 135.00  & 100.0 & 12.82 & \cmark & 45.75 & 9.77 & \cmark & 322 & 2 &  \\
\bottomrule
\end{tabular}
\label{tab:stat_outlier_sink}
\vspace{-10pt}
\end{table*}

%% file: tables/supp_sink_remove.tex
\begin{table}[t]
\centering
\setlength{\tabcolsep}{6pt}
\renewcommand{\arraystretch}{1.0}
\caption{\textbf{Effect of removing sink dimensions under $\Phi_{\text{LLM}}$ and $\Phi_{\text{VLM}}$ on AVHBench.} Removing the $\Phi_{\text{LLM}}$ sink token collapses performance, while removing tokens identified by $\Phi_{\text{VLM}}$ has little effect.}
\label{tab:sink_remove}
\resizebox{0.6\linewidth}{!}{
\begin{tabular}{l c c c c}
\toprule
\textbf{Method} & $A \to V$ & $V \to A$ & Matching & Overall \\
\midrule
Baseline & 82.31 & 82.58 & 51.71 & 71.60 \\
\midrule
All tokens & 0.00 & 0.00 & 0.00 & 0.00 \\
Sink token ($\Phi_{\text{LLM}}$) & 0.00 & 0.00 & 0.00 & 0.00 \\
Random token ($\Phi_{\text{VLM}}$) & 80.02 & 78.95 & 52.67 & 69.87 \\
Sink tokens ($\Phi_{\text{VLM}}$) & 78.96 & 82.14 & 52.77 & 71.06 \\
\bottomrule
\end{tabular}}
\label{tab:supp_sink_remove}
\end{table}

%% file: tables/vlm_tau_sensitive.tex
\begin{table*}[t]
\caption{\textbf{Sensitivity of $\Phi_{\mathrm{VLM}}$ to the sink threshold.}
We report the mean proportion of input tokens identified as sinks.
Increasing $\tau$ reduces the proportion of detected sinks, but the same threshold yields substantially different proportions across model scales.}
\vspace{-2mm}
\centering
\scriptsize
\setlength{\tabcolsep}{10pt}
\resizebox{\textwidth}{!}{%
\begin{tabular}{lccccc}
\toprule
\multirow{2}{*}{\textbf{Models}} &
\multirow{2}{*}{$\tau$} &
\multicolumn{4}{c}{\textbf{Sink Token Proportion (\%)}} \\
\cmidrule(lr){3-6}
& & OmniBench & AVUT & AVHBench & DailyOmni \\
\midrule
\multirow{3}{*}{Qwen2.5-Omni$_{3\mathrm{B}}$}
& 20 & 15.33 & 13.79 & 10.02 & 11.29 \\
& 25 &  2.50 &  1.91 &  1.30 &  1.38 \\
& 30 &  0.13 &  0.03 &  0.06 &  0.03 \\
\midrule
\multirow{3}{*}{Qwen2.5-Omni$_{7\mathrm{B}}$}
& 20 & 33.32 & 26.16 & 26.46 & 24.53 \\
& 25 & 16.41 & 11.83 & 12.09 & 11.44 \\
& 30 &  8.27 &  5.98 &  6.10 &  6.11 \\
\bottomrule
\end{tabular}
}
\vspace{-4mm}
\label{tab:sink_threshold}
\end{table*}

%% file: tables/supp_enhancement_sources.tex
\begin{table}[t]
\centering
\setlength{\tabcolsep}{6pt}
\renewcommand{\arraystretch}{1.0}
\caption{\textbf{Effect of different sink attention sources.} Using internal features achieves the best performance, whereas encoder inputs do not improve performance.}
\label{tab:enhancement_sources}
\resizebox{0.6\linewidth}{!}{
\begin{tabular}{l c c c c}
\toprule
\textbf{Mode} & $A \to V$ & $V \to A$ & Matching & Overall \\
\midrule
Baseline & 80.11 & 75.41 & 59.86 & 70.91 \\
\midrule
\rowcolor{gray!8}
Internal (Ours) & \textbf{80.37} & \textbf{75.81} & 60.18 & \textbf{71.25} \\
Encoder & 79.93 & 74.93 & \textbf{60.55} & 70.91 \\
Encoder w/o text & 80.11 & 75.33 & 59.59 & 70.78 \\
\bottomrule
\end{tabular}}
\end{table}

%% file: tables/rotation_vlm_top1.tex
\begin{table*}[t]
\caption{\textbf{Rotation with different sink identification criteria.}
We compare rotation toward sinks identified by $\Phi_{\mathrm{LLM}}$ with rotation toward the top 1\% or top 1 token under $\Phi_{\mathrm{VLM}}$.
All variants use rotation only, without mask relaxation.
We report mean accuracy and standard deviation over five seeds.
Restricting $\Phi_{\mathrm{VLM}}$ to fewer tokens does not recover an effective rotation direction.}
\vspace{-2mm}
\centering
\setlength{\tabcolsep}{10pt}
\resizebox{\textwidth}{!}{%
\begin{tabular}{llcccc}
\toprule
\multirow{2}{*}{\textbf{Models}} &
\multirow{2}{*}{\textbf{Method}} &
\multicolumn{4}{c}{\textbf{Audio-Visual QA}} \\
\cmidrule(lr){3-6}
& & OmniBench & AVUT & AVHBench & DailyOmni \\
\midrule
Qwen2.5-Omni$_{3\mathrm{B}}$
& Baseline
& $40.19 \pm 0.57$ & $60.23 \pm 0.56$ & $64.39 \pm 0.66$ & $55.03 \pm 0.70$ \\
\rowcolor{aliceblue}
& OutRo ($\Phi_{\mathrm{LLM}}$)
& $\mathbf{40.90 \pm 0.68}$ & $\mathbf{61.11 \pm 0.47}$ & $\mathbf{64.78 \pm 0.74}$ & $\mathbf{55.77 \pm 0.64}$ \\
& $\Phi_{\mathrm{VLM}}$ top 1\%
& $34.20 \pm 0.58$ & $48.69 \pm 0.50$ & $60.97 \pm 0.59$ & $33.02 \pm 0.49$ \\
& $\Phi_{\mathrm{VLM}}$ top 1
& $38.87 \pm 0.84$ & $56.38 \pm 1.01$ & $61.90 \pm 0.52$ & $43.97 \pm 0.54$ \\
\midrule
Qwen2.5-Omni$_{7\mathrm{B}}$
& Baseline
& $48.13 \pm 0.88$ & $\mathbf{64.53 \pm 0.53}$ & $71.13 \pm 0.32$ & $\mathbf{52.89 \pm 1.07}$ \\
\rowcolor{aliceblue}
& OutRo ($\Phi_{\mathrm{LLM}}$)
& $\mathbf{48.93 \pm 0.82}$ & $64.44 \pm 0.57$ & $\mathbf{72.03 \pm 0.27}$ & $52.70 \pm 1.16$ \\
& $\Phi_{\mathrm{VLM}}$ top 1\%
& $44.89 \pm 1.62$ & $63.32 \pm 1.16$ & $70.28 \pm 0.43$ & $49.40 \pm 1.24$ \\
& $\Phi_{\mathrm{VLM}}$ top 1
& $47.73 \pm 0.96$ & $63.70 \pm 0.47$ & $70.28 \pm 0.51$ & $51.06 \pm 1.37$ \\
\bottomrule
\end{tabular}
}
\vspace{-4mm}
\label{tab:sink_criterion_rotation}
\end{table*}

%% file: tables/image_text_llm.tex
\begin{table}[htbp]
\caption{\textbf{Generalisation to image--text and text-only settings.}
OutRo improves performance on POPE with image--text inputs and GSM8K with text-only inputs.
Results on Qwen2.5-Omni use greedy decoding over the full evaluation sets.
For text-only LLMs, we report mean accuracy and standard deviation over five seeds.}
\label{tab:image_text_generalisation}
\centering
\small
\setlength{\tabcolsep}{10pt}
\begin{tabular}{lcc}
\toprule
\multirow{2}{*}{\textbf{Models}} &
\textbf{Image--Text} & \textbf{Text-Only} \\
\cmidrule(lr){2-2} \cmidrule(lr){3-3}
& POPE & GSM8K \\
\midrule
Qwen2.5-Omni$_{3\mathrm{B}}$ & 87.67 & 73.92 \\
\rowcolor{aliceblue}
\textbf{\textit{\texttt{+}OutRo}} & \textbf{88.06} & \textbf{74.45} \\
\midrule
Qwen2.5-7B-Instruct & - & $59.03 \pm 0.73$ \\
\rowcolor{aliceblue}
\textbf{\textit{\texttt{+}OutRo}} & - & $\mathbf{61.41 \pm 0.85}$ \\
\midrule
Llama-3.1-8B-Instruct & - & $76.00 \pm 0.46$ \\
\rowcolor{aliceblue}
\textbf{\textit{\texttt{+}OutRo}} & - & $\mathbf{77.60 \pm 0.46}$ \\
\bottomrule
\end{tabular}
\end{table}

%% file: tables/random_rotation.tex
\begin{table*}[t]
\caption{\textbf{Rotation toward alternative directions.}
We compare the sink value direction with random video, audio, and text token directions and a head-mean direction on Qwen2.5-Omni 3B.
All variants use gated rotation without mask relaxation and preserve the original head-output norm.
We report mean accuracy and standard deviation over five seeds.}
\label{tab:random_direction_rotation}
\centering
\small
\setlength{\tabcolsep}{10pt}
\begin{tabular}{lcccc}
\toprule
\multirow{2}{*}{\textbf{Method}} &
\multicolumn{4}{c}{\textbf{Audio-Visual QA}} \\
\cmidrule(lr){2-5}
& OmniBench & AVUT & AVHBench & DailyOmni \\
\midrule
Baseline
& $40.19 \pm 0.57$ & $60.23 \pm 0.56$ & $64.39 \pm 0.66$ & $55.03 \pm 0.70$ \\
\rowcolor{aliceblue}
OutRo
& $\mathbf{40.90 \pm 0.68}$ & $\mathbf{61.11 \pm 0.47}$ & $\mathbf{64.78 \pm 0.74}$ & $\mathbf{55.77 \pm 0.64}$ \\
\midrule
Random video
& $36.38 \pm 1.28$ & $56.52 \pm 0.95$ & $61.04 \pm 0.39$ & $46.59 \pm 1.34$ \\
Random audio
& $37.14 \pm 0.70$ & $56.44 \pm 0.90$ & $60.57 \pm 0.92$ & $45.32 \pm 0.91$ \\
Random text
& $36.59 \pm 0.73$ & $53.42 \pm 0.99$ & $61.75 \pm 0.42$ & $44.49 \pm 1.55$ \\
Head mean
& $22.50 \pm 0.65$ & $26.18 \pm 0.70$ & $55.22 \pm 0.65$ & $25.98 \pm 0.75$ \\
\bottomrule
\end{tabular}
\end{table*}

%% file: algorithm/algorithm.tex
\begin{algorithm}[t]
\caption{OutRo in a Decoder Layer}
\label{alg:outro}
\begin{algorithmic}[1]
\Require Hidden states $\mathbf{X}^{(\ell)}$, sink criterion $\Phi_{\mathrm{LLM}}$, rotation strength $\gamma$
\Ensure Updated hidden states $\mathbf{X}^{(\ell)}_{\mathrm{out}}$

\State Identify sink indices:
\[
\mathcal{S}^{(\ell)} \leftarrow \Phi_{\mathrm{LLM}}(\mathbf{X}^{(\ell)})
\]

\State Compute attention outputs
\[
\mathbf{O}^{(\ell)}
=
\mathrm{Attn}(\mathbf{Q}^{(\ell)},\mathbf{K}^{(\ell)},\mathbf{V}^{(\ell)})
\]

\State Compute sink value direction
\[
\bar{\mathbf{v}}_{h,s}^{(\ell)}
=
\frac{1}{|\mathcal{S}^{(\ell)}|}
\sum_{s \in \mathcal{S}^{(\ell)}}
\mathbf{v}_{h,s}^{(\ell)}
\]

\Statex

\Comment{\textbf{Non-sink positions: gated head rotation}}
\For{each $ns \notin \mathcal{S}^{(\ell)}$, head $h$}

\State Compute directional alignment
\[
c_{h,ns}
=
\cos(\mathbf{O}_{h,ns}, \bar{\mathbf{v}}_{h,s}^{(\ell)})
\]

\State Compute gate
\[
g_{h,ns}
=
\tanh\!\left(
\frac{\mathrm{ReLU}(c_{h,ns})}{0.1}
\right)
\]

\State Rotate head output
\[
\hat{\mathbf{O}}_{h,ns}
=
\mathbf{O}_{h,ns}
+
\gamma g_{h,ns}
\frac{\mathbf{O}_{h,ns}\cdot\bar{\mathbf{v}}_{h,s}^{(\ell)}}
{\|\bar{\mathbf{v}}_{h,s}^{(\ell)}\|^2}
\bar{\mathbf{v}}_{h,s}^{(\ell)}
\]

\State Rescale magnitude
\[
\mathbf{O}_{h,ns}
\leftarrow
\frac{\|\mathbf{O}_{h,ns}\|}
{\|\hat{\mathbf{O}}_{h,ns}\|}
\hat{\mathbf{O}}_{h,ns}
\]

\EndFor

\Statex

\Comment{\textbf{Sink positions: mask relaxation}}
\If{$\ell = \ell_{\mathrm{enh}}$}
\For{each $s \in \mathcal{S}^{(\ell)}$, head $h$}

\State Relax causal mask for sink queries
\[
\mathbf{A}^{\mathrm{relaxed}}_{h,s,:}
=
\mathrm{Softmax}
\!\left(
\frac{\mathbf{Q}_{h,s}\mathbf{K}_h^\top}{\sqrt{D_h}}
\right)
\]

\State Replace sink-position outputs
\[
\mathbf{O}^{(\ell)}_{h,s}
\leftarrow
\mathbf{A}^{\mathrm{relaxed}}_{h,s,:}\mathbf{V}_h
\]

\EndFor
\EndIf

\State Aggregate heads and apply output projection
\[
\mathbf{X}^{(\ell)}_{\mathrm{out}}
\leftarrow
\mathbf{X}^{(\ell)} + W_O \mathbf{O}^{(\ell)}
\]

\State \Return $\mathbf{X}^{(\ell)}_{\mathrm{out}}$

\end{algorithmic}
\end{algorithm}